\pdfoutput=1

\documentclass[11pt]{article}

\usepackage[final]{acl}

\usepackage{times}
\usepackage{latexsym}

\usepackage[T1]{fontenc}

\usepackage[utf8]{inputenc}

\usepackage{microtype}

\usepackage{inconsolata}

\usepackage{booktabs}
\usepackage{multirow}
\usepackage{multicol}
\usepackage{makecell}
\usepackage{enumitem}
\usepackage{balance}
\usepackage{graphicx}
\usepackage{subcaption}
\usepackage{amsmath}
\usepackage{amsthm}
\usepackage{amsfonts}
\usepackage{hyperref}
\definecolor{lightred}{rgb}{0.8,0.85,1}
\definecolor{lightred}{rgb}{1, 0.8, 0.8}
\definecolor{darkgreen}{RGB}{0,150,0}
\usepackage[ruled,vlined]{algorithm2e}
\usepackage[breakable,skins]{tcolorbox}
\usepackage{alltt}
\newtcolorbox{AIBox}[2][]{aibox,title=#2,#1}
\newcommand{\mybox}[4]{
    \begin{figure}[h]
        \centering
    \begin{tikzpicture}
        \node[anchor=text,text width=\columnwidth-0.45cm, draw, rounded corners, line width=1pt, fill=#3, inner sep=2.5mm, align=justify] (big) {\\#4};
        \node[draw, rounded corners, line width=1pt, fill=#2, anchor=west, xshift=5mm] (small) at (big.north west) {#1};
    \end{tikzpicture}
    \end{figure}
}
\tcbset{
  aibox/.style={
    width=0.95\textwidth,
    top=5pt,
    colback=black!05,
    colframe=black!20,
    colbacktitle=black!50,
    enhanced,
    center,
    attach boxed title to top left={yshift=-0.1in,xshift=0.1in},
    boxed title style={boxrule=0pt,colframe=white,},
  }
}

\title{Unveiling the Key Factors for Distilling Chain-of-Thought Reasoning}

\author{
  \textbf{Xinghao Chen$^{1,2,3}$}, 
  \textbf{Zhijing Sun$^{3}$}, 
  \textbf{Wenjin Guo$^{3}$}, 
  \textbf{Miaoran Zhang$^{4}$}, 
  \textbf{Yanjun Chen$^{1,3}$}, \\
  \textbf{Yirong Sun$^{3}$}, 
  \textbf{Hui Su$^{5}$}, 
  \textbf{Yijie Pan$^{3}$}, 
  \textbf{Dietrich Klakow$^{4}$}, 
  \textbf{Wenjie Li$^{1}$}\thanks{Corresponding authors.}, 
  \textbf{Xiaoyu Shen$^{2,3}$}\footnotemark[1] \\
  $^1$ Department of Computing, The Hong Kong Polytechnic University\\
  $^2$ Ningbo Key Laboratory of Spatial Intelligence and Digital Derivative\\ $^3$Institute of Digital Twin, EIT \qquad
  $^4$ Saarland University \qquad $^5$ Meituan Inc.\\
  \texttt{
  xing-hao.chen@connect.polyu.hk} \\
  \texttt{cswjli@comp.polyu.edu.hk
  \qquad xyshen@eitech.edu.cn}
}
\begin{document}
\maketitle
\begin{abstract}
Large Language Models (LLMs) excel in reasoning tasks through Chain-of-Thought (CoT) prompting.
However, CoT prompting greatly increases computational demands, which has prompted growing interest in distilling CoT capabilities into Small Language Models (SLMs). This study systematically examines the factors influencing CoT distillation,  including the choice of granularity, format and teacher model. Through experiments involving four teacher models and seven student models across seven mathematical and commonsense reasoning datasets, we uncover three key findings: (1) Unlike LLMs, SLMs exhibit a \emph{non-monotonic} relationship with granularity, with stronger models benefiting from finer-grained reasoning and weaker models performing better with simpler CoT supervision; (2) CoT format significantly impacts LLMs but has \emph{minimal} effect on SLMs, likely due to their reliance on supervised fine-tuning rather than pretraining preferences; (3) Stronger teacher models do \emph{NOT} always produce better student models, as diversity and complexity in CoT supervision can outweigh accuracy alone. These findings emphasize the need to tailor CoT strategies to specific student model, offering actionable insights for optimizing CoT distillation in SLMs. The code and datasets are available at
\href{https://github.com/EIT-NLP/Distilling-CoT-Reasoning}{\nolinkurl{https://github.com/EIT-NLP/Distilling-CoT-Reasoning}}.
\end{abstract}

\section{Introduction}
\begin{figure}[t!]
\raggedright
\includegraphics[width=\linewidth]{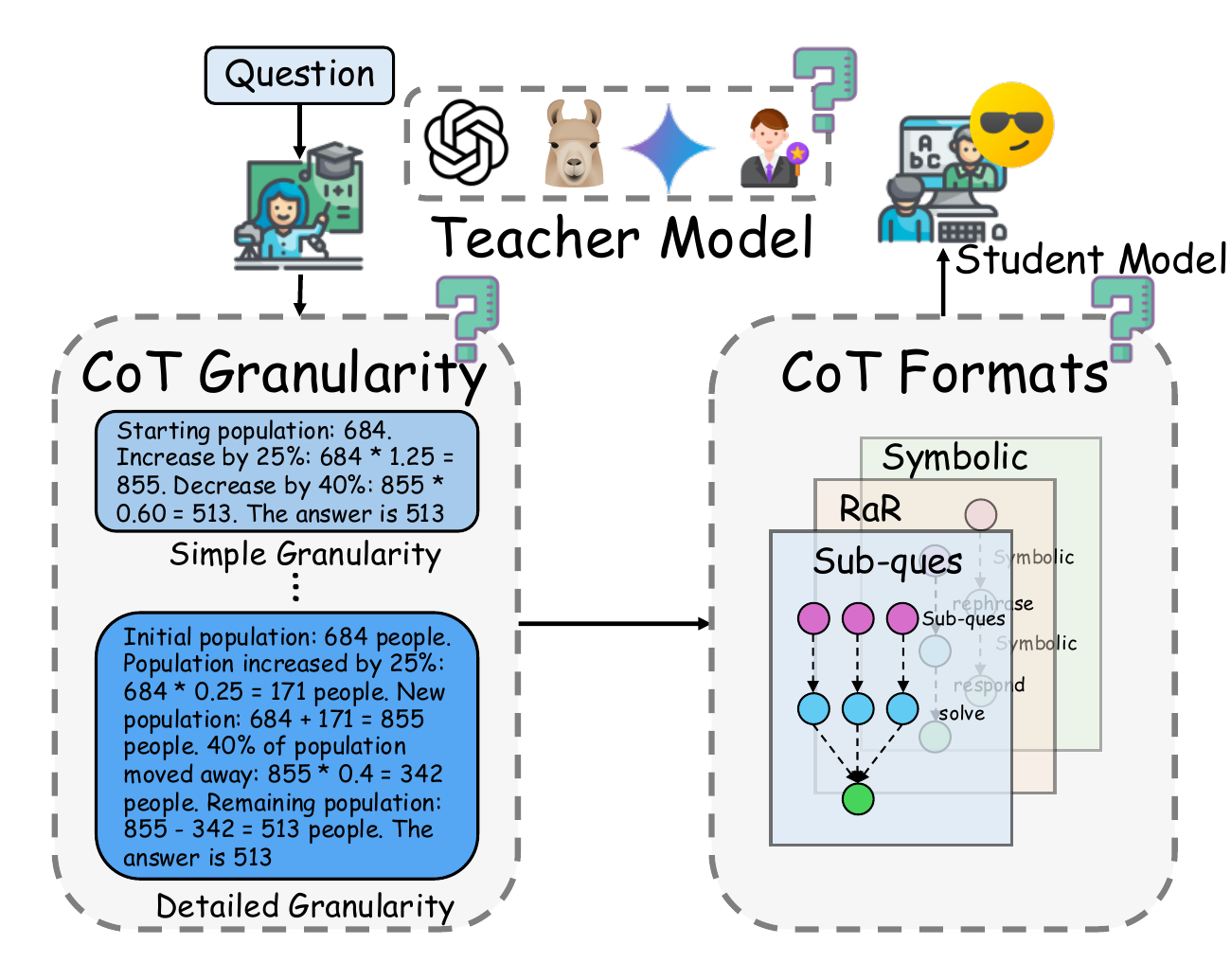}
\caption{\small Overview of CoT Distillation. Different teacher models generate CoT supervision with varying levels of granularity and formats to fine-tune the student model.}
\label{fig:granularity}
\end{figure}

Large Language Models (LLMs) have demonstrated exceptional capabilities through extensive pretraining on diverse human language data~\cite{brown2020language, hoffmann2022training, team2024gemini, meta2024llama3, openai2024gpt4o, openaio1}. Chain-of-Thought (CoT) prompting has further enhanced their abilities by guiding LLMs to generate intermediate reasoning tokens, which emulate human cognitive processes and improve interpretability~\cite{kojima2022large, wei2023chainofthoughtpromptingelicitsreasoning}. Advances in CoT prompting have explored techniques like extending reasoning steps~\cite{jin-etal-2024-impact, merrill2024the} and refining reasoning formats~\cite{
deng2024rephraserespondletlarge, xu2024faithfullogicalreasoningsymbolic}. However, CoT's token-intensive nature significantly increases computational demands~\cite{zhao2024surveylargelanguagemodels}, limiting its practicality in resource-constrained settings. This has spurred interest in distilling CoT capabilities into Small Language Models (SLMs) as a more efficient alternative~\cite{team2024gemma, meta2024Llama32}.

Since SLMs often struggle to independently generate effective CoT reasoning solutions~\cite{kaplan2020scalinglawsneurallanguage, stolfo-etal-2023-causal}, distilling CoT capabilities requires fine-tuning SLMs on teacher-annotated CoT datasets, where the teacher can be either human experts or more powerful LLMs. While previous research has demonstrated successful distillation of CoT capabilities into SLMs~\cite{ho-etal-2023-large, magister-etal-2023-teaching, xu2024surveyknowledgedistillationlarge,deepseekai2025deepseekr1incentivizingreasoningcapability}, the choice of teacher annotators and CoT generation methods has often been arbitrary. A critical, yet unexplored, research question remains: \emph{What is the most effective CoT supervision for training a student model to develop robust reasoning capabilities?} Analogous to how human teachers instruct students, there are three key factors that influence how effectively a student absorbs knowledge:
\begin{itemize}
    \item \textbf{Choice of teacher}: This defines \emph{who} teaches the student. Different teachers bring varying levels of knowledge, teaching styles, and problem-solving approaches. In reality a student’s performance can vary significantly depending on the teacher, and the most knowledgeable person is not always the best teacher. 
    \item \textbf{Granularity of teaching}: This defines \emph{what} level of detail is provided. Teachers may provide varying levels of explanation: some offer detailed, step-by-step reasoning, while others skip over simpler steps, assuming they are self-evident. The optimal level of granularity depends on the student’s perspective of what needs to be explained. 
    
    \item \textbf{Format of teaching}: This defines \emph{how} the reasoning is structured and presented. Even with the same teacher and granularity level, the way information is organized and expressed can significantly impact learning outcomes. Some students may prefer plain language explanations, while others may thrive with more technical, mathematical language.
\end{itemize}
Building on this analogy of how human teaching impacts student performance, we conducted extensive experiments on four mathematical reasoning datasets of varying difficulty and three commonsense reasoning datasets, using four teacher models to distill reasoning skills to seven student models. We adopted a 1-shot prompting approach for generating CoT annotations, which we found to be the most effective in maintaining consistency in teaching style while controlling granularity.
Our key findings are: (1) While LLMs benefit monotonically from detailed steps, SLMs exhibit an \emph{non-monotonic relationship}. Stronger student models benefit from finer granularity, while weaker ones can be overwhelmed by excessive explanations and prefer simpler CoT annotations; (2) CoT format changes influence LLMs, likely due to their pretraining preferences for certain structures, but this effect is \emph{less pronounced} in SLMs, which adapt more readily to diverse formats during fine-tuning; (3) Contrary to prior research suggesting that better teacher models invariably lead to better student performance~\cite{zong2023better}, in the task of distilling CoT capabilities, we find that \emph{better teacher models do not always produce better student models}. Sronger student models benefit more
from advanced teacher model. Human-annotated CoTs, despite their near-perfect accuracy, often underperform LLM-generated CoTs. Our work presents the first systematic framework for optimizing CoT distillation, laying the groundwork for enhancing the reasoning capabilities of SLMs.

\section{Related Work}

\paragraph{CoT prompting} CoT prompting~\cite{wei2023chainofthoughtpromptingelicitsreasoning} has become a pivotal technique for enhancing reasoning capabilities in LLMs by introducing intermediate reasoning steps. Automated approaches like Auto-CoT~\cite{zhang2023automatic}, Tree-of-Thoughts~\cite{yao2023tree} and Self-play Mutual Reasoning~\cite{qi2024mutualreasoningmakessmaller} explore multiple reasoning paths to expand the search space and improve task accuracy. These methods focus on increasing the reasoning length or expanding the reasoning horizon to handle complex tasks. 
Recent studies have underscored the importance of reasoning granularity and formats in enhancing LLM performance. For instance, \citet{jin-etal-2024-impact} identified that longer reasoning steps improve task success for complex problems, while overly concise steps can reduce effectiveness. Tailored reasoning formats\cite{DBLP:conf/iclr/KhotTFF0CS23, zhou2023leasttomost, deng2024rephraserespondletlarge, xu2024faithfullogicalreasoningsymbolic} have demonstrated substantial improvements across tasks. However, these reasoning optimization strategies often comes with significant computational costs~\cite{nayab2024concisethoughtsimpactoutput}, raising concerns about the trade-off between accuracy and efficiency. 

\paragraph{Knowledge distillation} While direct prompting enables LLMs to perform complex reasoning through CoT, SLMs struggle due to limited capacity~\cite{stolfo-etal-2023-causal}. Knowledge distillation (KD) provides an effective framework for transferring the reasoning capabilities of teachers to SLMs~\cite{xu2024surveyknowledgedistillationlarge}. A simple yet effective approach is using a teacher-student paradigm, which employs teacher-generated CoT steps to guide SLMs, addressing their limitations and enhancing reasoning-intensive task performance~\cite{magister-etal-2023-teaching, ho-etal-2023-large, shridhar-etal-2023-distilling}. 
Despite these advances, a systematic exploration of how to balance reasoning granularity, format, and teaching strategies remains lacking. Addressing these gaps is crucial for optimizing CoT distillation and enabling efficient reasoning in SLMs.

\section{Problem Formulation}
\label{sec:preliminaries}
Let $\mathcal{D}=\{(x_i, y_i)\}_{1}^{N}$ denote a reasoning dataset with $N$ $(x_i, y_i)$ pairs. Chain-of-Thought distillation aims to train a student $S$ to generate intermediate reasoning steps $C_i$ for each input $x_i$ in order to generate the right $y_i$. The optimal $C_i$ to train the student model is influenced by three key pedagogical factors: \textbf{Choice of Teacher}, \textbf{Granularity of Teaching}, and \textbf{Format of Teaching}:

\paragraph{Choice of Teacher}

The teacher model \( T \) generates a reasoning chain \( \mathcal{C}_T(x_i) \) for each input \( x_i \), which guides the student model in producing the correct answer. The teacher can either be an LLM or a human with varying styles and expertise.

\paragraph{Granularity of Teaching}
Granularity refers to the level of detail in the CoT reasoning. A high-granularity annotation provides detailed, step-by-step reasoning, while a low-granularity annotation skips steps and provides a more abstract summary. We represent the CoT chain with granularity \( g \) as
\(
\mathcal{C}_g(x_i) = (c_{g,1}, c_{g,2}, \dots, c_{g,k_g})
\)
where \( k_g \) is the number of reasoning steps. Higher $k_g$ and more tokens in $c_{g, i}$ indicates higher granularity.

\paragraph{Format of Teaching}

Format refers to the structure in which the CoT reasoning is presented. It could be in natural language, formal logic, or symbolic representation. We denote the CoT chain in format \( f \) as \(
\mathcal{C}_f(x_i)\). The format impacts how the reasoning steps are conveyed.

Given these three factors, the distillation process involves supervised fine-tuning of the student model \( S \) on generated CoT annotations:
\[
\mathcal{L}_{\text{distill}} = \sum_{i=1}^{N} \mathcal{L}(S(x_i), \mathcal{C}_{T,g,f}(x_i) \oplus y_i)
\]
where $S(x_i)$ is the generation from $S$, $\mathcal{C}_{T,g,f}(x_i)$ denotes the CoT annotation generated under teacher $T$ with granularity $g$ and format $f$, $\oplus$ denotes concatenation and \( \mathcal{L} \) measures the discrepancy between $S(x_i)$ and the ground truth.

\section{Experimental setup}

\begin{figure*}[t]
    \centering
    \begin{subfigure}[t]{0.48\textwidth}
        \centering
        \includegraphics[width=\textwidth]{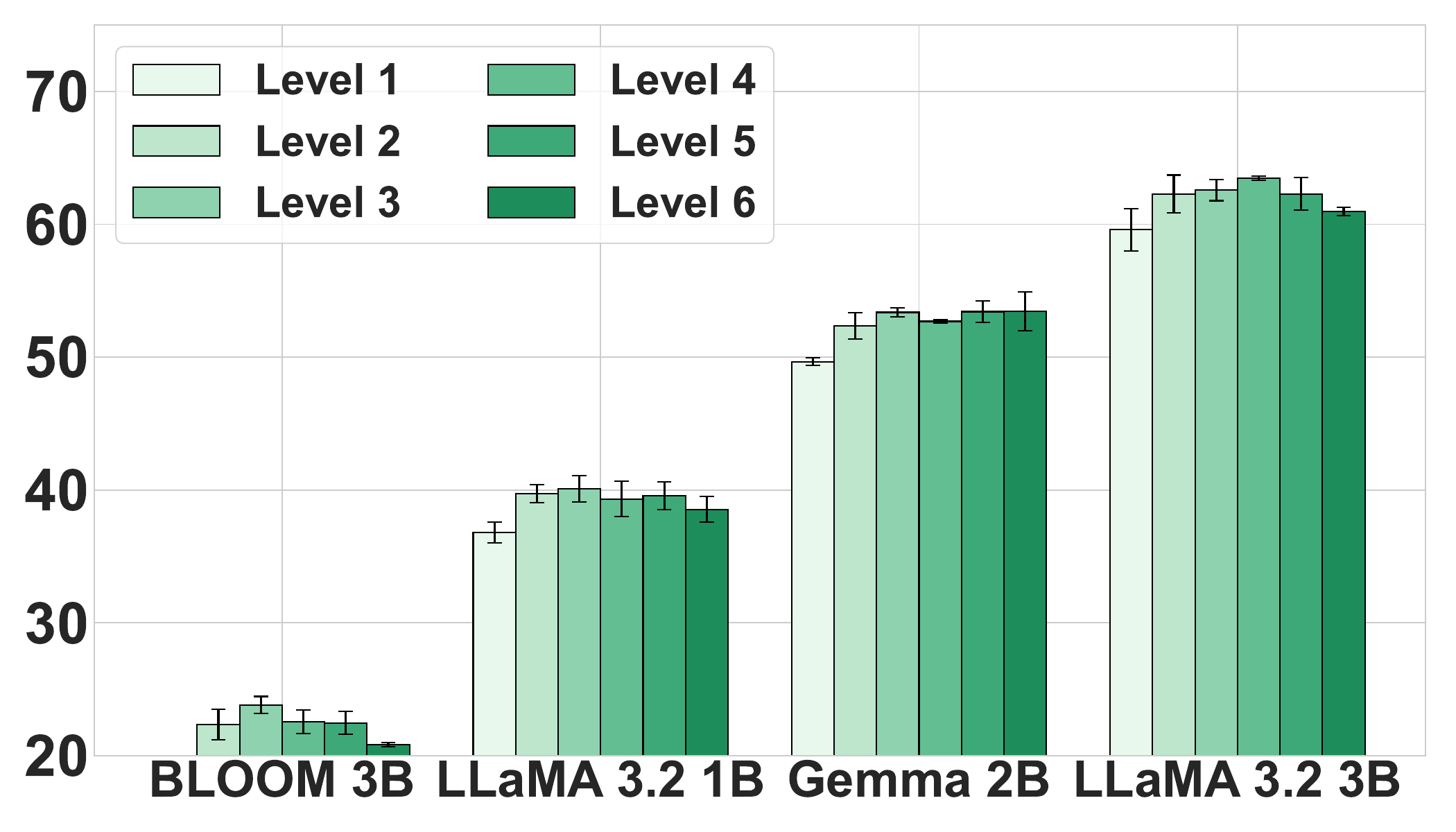}
        \caption{GSM8K}
    \end{subfigure}%
    \hfill
    \begin{subfigure}[t]{0.48\textwidth}
        \centering
        \includegraphics[width=\textwidth]{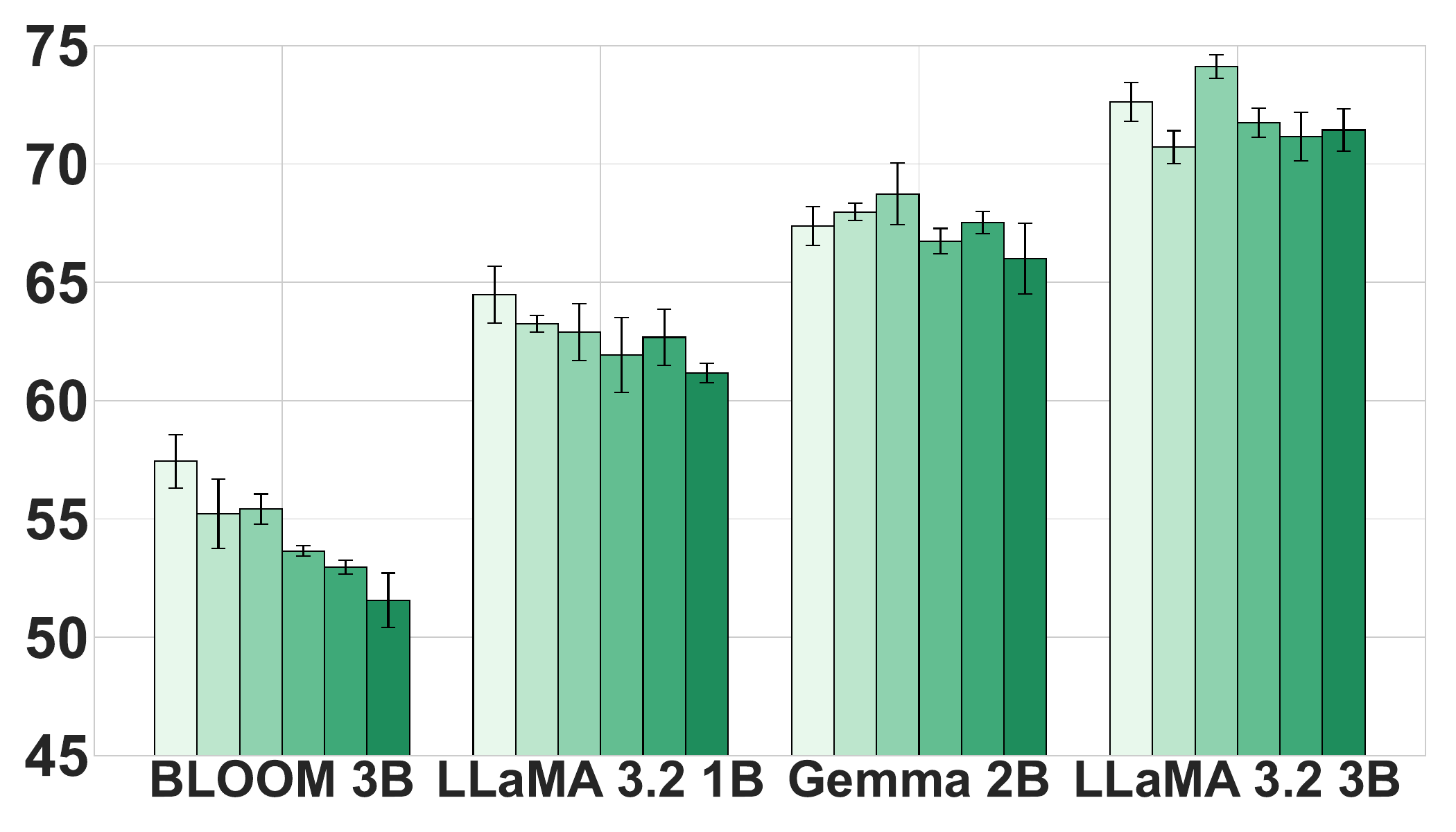}
        \caption{CommonsenseQA}
    \end{subfigure}
    \caption{Performance of student models with different granularity. Most models achieve peak accuracy at intermediate granularity levels.}
    \label{fig:combined_granularity}
\end{figure*}

\subsection{Generation of CoT Annotation}

\paragraph{Teacher Models}
We use three teacher models: \textbf{GPT-4o}~\cite{openai2024gpt4o}, \textbf{Gemini-1.5-Flash}~\cite{team2024gemini}, and \textbf{LLaMA 3 70B}~\cite{meta2024llama3}, chosen for their diverse architectures and reasoning capacities. Additionally, we include \textbf{human-annotated} CoTs, typically considered the ground-truth reasoning steps~\cite{kumar2024enhancing}. 

\paragraph{Generation Method}
The CoT generation process begins with selecting a representative problem from the training split as a 1-shot example. This example is used to prompt teacher models for generating annotations under various configurations. For granularity, we prompt the model to generate CoTs with varying levels of detail simultaneously. For format, we prompt the model to generate CoTs for each format individually.~\footnote{For granularity, we also investigated other data collection strategies, such as generating reasoning steps forward and backward simultaneously, but these methods did not produce better data, as shown in Appendix~\ref{appendix:Granularity_workflow}.}
Details regarding the workflow, prompt designs, and case studies are included respectively in Appendix~\ref{appendix:Granularity} and~\ref{appendix:Format}.

\subsection{Tasks and Datasets}
\paragraph{Mathematical Reasoning}
To evaluate mathematical reasoning, we utilize four datasets with varying complexity levels:
\textbf{SVAMP}~\cite{patel2021nlpmodelsreallyable}, 
\textbf{GSM8K}~\cite{cobbe2021trainingverifierssolvemath}, 
\textbf{AQuA-RAT}~\cite{ling-etal-2017-program}, 
\textbf{MATH}~\cite{hendrycksmath2021}. 
SVAMP, GSM8K, and MATH require numerical answers, while AQuA-RAT adopts a multiple-choice format. From the MATH dataset, we randomly sample problems from subcategories such as prealgebra, algebra, number theory, and counting and probability, ensuring a representative coverage of diverse mathematical domains. 

\paragraph{Commonsense Reasoning}
For commonsense reasoning, we use three datasets:
\textbf{CommonsenseQA} (CSQA, \citealt{talmor-etal-2019-commonsenseqa, aggarwal-etal-2021-explanations}), \textbf{OpenBookQA} (OBQA, \citealt{mihaylov2018suitarmorconductelectricity}), and \textbf{StrategyQA} (STQA,  \citealt{geva2021didaristotleuselaptop}). These datasets test the models’ ability to handle everyday reasoning and general knowledge tasks. CSQA uses a 5-class multiple-choice format, OBQA has 4 classes, and STQA is binary. 

For evaluation, answers are extracted from generated responses using predefined templates and regular expressions. We use accuracy as our evaluation metric, which is calculated as the ratio of correctly predicted instances to the total number of instances: $\text{Accuracy} = N_{\text{correct}} / N_{\text{total}}$. The complete details can be found in the Appendix~\ref{appendix:dataset}.

\section{Effects of Granularity}
\label{sec:granularity}

While previous research has shown that increasing reasoning granularity improves LLM performance through detailed intermediate steps~\cite{jin-etal-2024-impact, merrill2024the}, SLMs differ fundamentally from LLMs in their ability to process complex reasoning chains. This raises a critical question: \emph{does increasing reasoning granularity still yield consistent benefits for SLMs in the task of CoT distillation?} In this section, we investigate this question using GPT-4o as the teacher model.

\paragraph{Non-Monotonic Scaling in Student Models}

\begin{table*}[ht]
\centering
\small
\renewcommand{\arraystretch}{0.6}
\resizebox{1.0\linewidth}{!}{
\begin{tabular}{>{\centering\arraybackslash}m{2.5cm} |c|llllll}
\toprule
\toprule
\multirow{2}{*}{\textbf{Dataset}} & \multirow{2}{*}{\textbf{Only Answer}} & \multicolumn{6}{c}{\textbf{Gemma 2B Performance}} \\ 
\cmidrule(lr){3-8}
& & Level 1 & Level 2 & Level 3 & Level 4 & Level 5 & Level 6 \\
\midrule
SVAMP & 47.70 & 59\textsubscript{±4.58} & 64.33\textsubscript{±0.00} & 65.22\textsubscript{±0.69} & 65.89\textsubscript{±0.38} & \textbf{67.11\textsubscript{±1.35}}$^{\color{red}{\uparrow\scriptsize13.74\%}}$ & 66.89\textsubscript{±1.02} \\
GSM8K & 8.20 & 49.66\textsubscript{±0.27} & 52.36\textsubscript{±0.98} & 53.37\textsubscript{±0.33} & 52.69\textsubscript{±0.13} & 53.42\textsubscript{±0.83}& \textbf{53.45\textsubscript{±1.48}$^{\color{red}{\uparrow\scriptsize7.63\%}}$ } \\
AQuA-RAT & 20.47 & 40.68\textsubscript{\textit{±1.27}} & 42.91\textsubscript{±1.42} & 43.7\textsubscript{±2.58} & 39.9\textsubscript{±1.49} & \textbf{44.88}\textsubscript{±0.79}$^{\color{red}{\uparrow\scriptsize12.48\%}}$ & 44.49\textsubscript{±2.36} \\
MATH & 9.00 & 23.4\textsubscript{±1.06} & 21.53\textsubscript{±2.16} & \textbf{24.4\textsubscript{±0.20}}$^{\color{red}{\uparrow\scriptsize16.19\%}}$ & 21.93\textsubscript{±0.42} & 23.0\textsubscript{±1.22} & 21.0\textsubscript{±0.69} \\
CSQA & \textbf{69.86} & 67.38\textsubscript{±0.82} & 67.98\textsubscript{±0.37} & 
68.74\textsubscript{±1.30} & 
66.75\textsubscript{±0.53} & 67.54\textsubscript{±0.47} & 66.01\textsubscript{±1.50} \\
OBQA & 69.60 & 71.53\textsubscript{±1.94} & 69.93\textsubscript{±0.90} & 69.93\textsubscript{±1.36}& 68.33\textsubscript{±1.27} & \textbf{72.00\textsubscript{±1.64}}$^{\color{red}{\uparrow\scriptsize5.37\%}}$ & 70.13\textsubscript{±1.62} \\
STQA & 60.69 & \textbf{67.59\textsubscript{±1.04}}$^{\color{red}{\uparrow\scriptsize7.11\%}}$ & 63.1\textsubscript{±1.79} & 64.6\textsubscript{±1.56} & 63.45\textsubscript{±1.24} & 65.75\textsubscript{±1.77} & 64.14\textsubscript{±1.58} \\
\midrule
\multirow{2}{*}{\textbf{Dataset}} & \multirow{2}{*}{\textbf{Only Answer}} & \multicolumn{6}{c}{\textbf{LLaMA 3.2 3B Performance}} \\ 
\cmidrule(lr){3-8}
& & Level 1 & Level 2 & Level 3 & Level 4 & Level 5 & Level 6 \\
\midrule
SVAMP & 53.70 & 69\textsubscript{±3.61} & 65.89\textsubscript{±3.53} & 68.33\textsubscript{±1.86} & 68.11\textsubscript{±3.27} & 69.78\textsubscript{±1.07} & \textbf{74.33\textsubscript{±1.45}}$^{\color{red}{\uparrow\scriptsize12.81\%}}$ \\ 
GSM8K & 9.30 & 59.59\textsubscript{±1.59} & 62.29\textsubscript{±1.41} & 62.57\textsubscript{±0.80} & \textbf{63.48\textsubscript{±0.16}}$^{\color{red}{\uparrow\scriptsize6.53\%}}$ & 62.29\textsubscript{±1.22} & 60.98\textsubscript{±0.31} \\ 
AQuA-RAT & 19.60 & 44.36\textsubscript{±2.31} & 44.88\textsubscript{±2.19} & 45.01\textsubscript{±2.37}& 46.19\textsubscript{±3.94} & \textbf{47.24\textsubscript{±4.77}}$^{\color{red}{\uparrow\scriptsize6.49\%}}$ & 46.33\textsubscript{±3.01} \\ 
MATH & 9.40 & 19.07\textsubscript{±0.90} & 19.6\textsubscript{±1.06} & 19.73\textsubscript{±1.72} & \textbf{20.27\textsubscript{±1.42}}$^{\color{red}{\uparrow\scriptsize11.37\%}}$ & 19.93\textsubscript{±2.20} & 18.2\textsubscript{±1.64} \\ 
CSQA & 62.00 & 72.62\textsubscript{±0.82} & 70.71\textsubscript{±0.70} & \textbf{74.12\textsubscript{±0.50}}$^{\color{red}{\uparrow\scriptsize4.82\%}}$ & 71.75\textsubscript{±0.62} & 71.17\textsubscript{±1.03} & 71.44\textsubscript{±0.90} \\ 
OBQA & 74.40 & 79.33\textsubscript{±0.42} & 79.73\textsubscript{±0.70} & 78.8\textsubscript{±0.80} & 77.8\textsubscript{±0.92} & 79.27\textsubscript{±2.04} & \textbf{80.2\textsubscript{±1.78}}$^{\color{red}{\uparrow\scriptsize3.08\%}}$ \\ 
STQA & 55.52 & 66.44\textsubscript{±1.55} & 62.76\textsubscript{±2.82} & 67.47\textsubscript{±1.44} & 66.78\textsubscript{±1.39} & 63.91\textsubscript{±2.79} & \textbf{68.62\textsubscript{±1.20}}$^{\color{red}{\uparrow\scriptsize9.34\%}}$ \\ 
\midrule
\multirow{2}{*}{\textbf{Dataset}} & \multirow{2}{*}{\textbf{Only Answer}} & \multicolumn{6}{c}{\textbf{BLOOM 3B Performance}} \\ 
\cmidrule(lr){3-8}
& & Level 1 & Level 2 & Level 3 & Level 4 & Level 5 & Level 6 \\
\midrule
SVAMP & 5.00 & 15.44\textsubscript{±0.51} & 23.67\textsubscript{±0.00} & 23.11\textsubscript{±1.26} & \textbf{24.00\textsubscript{±0.67}}$^{\color{red}{\uparrow\scriptsize55.44\%}}$ & 22.22\textsubscript{±0.69} & 22.22\textsubscript{±1.02} \\ 
GSM8K & 4.60 & 18.2\textsubscript{±0.57} & 22.34\textsubscript{±1.14} & \textbf{23.81\textsubscript{±0.65}}$^{\color{red}{\uparrow\scriptsize30.82\%}}$ & 22.57\textsubscript{±0.88} & 22.47\textsubscript{±0.86} & 20.85\textsubscript{±0.15} \\ 
AQuA-RAT & \textbf{28.00}
& 24.67\textsubscript{±0.82} & 24.41\textsubscript{±1.72} & 20.34\textsubscript{±1.82} & 26.90\textsubscript{±2.41} & 25.85\textsubscript{±0.45} & 24.28\textsubscript{±2.17} \\ 
MATH & \textbf{4.60} & 3.2\textsubscript{±1.04} & 2.8\textsubscript{±0.40} & 2.33\textsubscript{±0.61} & 2.73\textsubscript{±0.23} & 3.53\textsubscript{±0.50} & 2.8\textsubscript{±0.20} \\ 
CSQA & 20.56 & \textbf{57.44\textsubscript{±1.12}}$^{\color{red}{\uparrow\scriptsize11.38\%}}$ & 55.23\textsubscript{±1.47} & 55.42\textsubscript{±0.64} & 53.65\textsubscript{±0.22} & 52.96\textsubscript{±0.29} & 51.57\textsubscript{±1.15} \\ 
OBQA & 37.80 & 57.2\textsubscript{±1.59} & 52.33\textsubscript{±0.61} & 54.87\textsubscript{±2.02} & 54.6\textsubscript{±1.04} & \textbf{57.47\textsubscript{±2.64}}$^{\color{red}{\uparrow\scriptsize9.82\%}}$ & 52.93\textsubscript{±1.81} \\ 
STQA & 54.14 & 58.85\textsubscript{±1.74} & \textbf{61.04\textsubscript{±3.06}}$^{\color{red}{\uparrow\scriptsize3.72\%}}$ & 60.58\textsubscript{±3.09} & 59.89\textsubscript{±3.13} & 59.19\textsubscript{±1.21} & 59.08\textsubscript{±2.87} \\ 
\bottomrule
\bottomrule
\end{tabular}
}
\caption{Performance of Gemma 2B, LLaMA 3.2 3B and BLOOM 3B at six granularity levels. For each dataset, the best performance is boldfaced, and red text shows the relative improvement (\%) for highest vs. lowest performance in six levels. \textit{Only Answer}: Student models are fine-tuned to directly predict answers without CoT.}
\label{tab:granularity_performance}
\end{table*}

As shown in Figure~\ref{fig:granularity}, our experiments reveal a \emph{non-monotonic relationship between CoT granularity and student model accuracy}. 
Most models exhibit peak performance at intermediate granularity levels. Further increasing granularity leads to diminishing returns and even performance declines. 
It suggests that intermediate granularity strikes a balance between informativeness and efficiency in CoT, whereas overly detailed reasoning chains may introduce redundant information which is overwhelming especially for weaker models.

Table \ref{tab:granularity_performance} presents the performance of three representative student models across seven evaluation datasets. We include a baseline called \textit{Only Answer}, where student models are fine-tuned to predict answers without CoT. Similar to System 1’s automatic thinking \cite{yu2024distilling21}, higher baseline score suggests that the model may have implicitly learned the relevant knowledge during pretraining \cite{prabhakar-etal-2024-deciphering}.

Notably, \emph{stronger and more recent student models}, such as those from the Gemma and LLaMA family, achieve significant performance gains from KD at \emph{higher granularity levels}. In contrast, \emph{smaller and weaker models} like BLOOM family improve on simpler tasks such \emph{at the intermediate granularity levels} but struggle on more challenging datasets, sometimes performing no better than random guessing. This trend of BLOOM family aligns with parameter scaling laws \cite{kaplan2020scalinglawsneurallanguage} for simpler tasks but breaks down for more complex ones, where smaller models fail to acquire the reasoning abilities due to limited training data. Full results are provided in Appendix \ref{appendix:Granularityresults}.

These findings emphasize that CoT granularity plays a crucial role in CoT distillation.
Customizing granularity levels to align with the student’s abilities is thus critical for maximizing the efficiency and effectiveness.

\paragraph{Distinguishing Granularity from Length Effect}
\label{sec:length}
Increasing reasoning granularity often leads to longer sequences as a byproduct. To isolate the impact of granularity from sequence length, we pad CoT training samples for a lower granularity level $g_1$ with non-informative filler content to match the sequence length of a higher granularity $g_2$, such that $\text{avg\_len}(\mathcal{D}_{g_1}) \approx \text{avg\_len}(\mathcal{D}_{g_2})$. This modification allows us to assess whether reasoning accuracy stems from granularity or simply sequence length. The specific padding procedure can be found in the Appendix~\ref{appendix:Padding}.

As shown in Table \ref{tab:length_effects}, padding Level 1 reasoning chains to level 5 consistently failed to replicate the gains observed with actual higher-granularity reasoning, which demonstrates that \emph{simply increasing sequence length without introducing meaningful reasoning steps does not enhance model performance}. Furthermore, adding filler content may introduce noise or distract the model, leading to degraded performance \cite{zhou2024can, li2024happenedllmslayerstrained}. This highlights the critical role of granularity, rather than sequence length alone, in driving reasoning efficacy. 

\begin{table}[ht]
\centering
\small
\resizebox{\columnwidth}{!}{
\begin{tabular}{lcccc}
\toprule
\multirow{2}{*}{\textbf{Granularity}} & \multicolumn{2}{c}{\textbf{GSM8K}} & \multicolumn{2}{c}{\textbf{AQuA-RAT}} \\ 
\cmidrule(lr){2-3} \cmidrule(lr){4-5}
 & \textbf{Acc} & \textbf{Seq. Length} & \textbf{Acc} & \textbf{Seq. Length} \\ 
\midrule
 Level 1 & 47.61 & 100.93 & 40.15 & 149.31 \\
 Level 1 Padded & 46.62 & 143.43 & 37.80 & 220.34 \\
 Level 5 & \textbf{52.92} & 138.16 & \textbf{42.51} & 216.13 \\
\bottomrule
\end{tabular}
}
\caption{Performance and sequence length of Gemma 2B on GSM8k and AQuA-RAT with varying granularity levels and padding conditions.}
\label{tab:length_effects}
\end{table}

\begin{table*}[ht]
\centering
\resizebox{1.0\textwidth}{!}{
\begin{tabular}{c|c|c|c|c|c|c|c|c}
\toprule
\toprule
\textbf{Dataset} & \textbf{CoT Format} & \textbf{BLOOM 560M} & \textbf{BLOOM 1.1B} & \textbf{BLOOM 1.7B} & \textbf{BLOOM 3B} & \textbf{Gemma 2B} & \textbf{LLaMA 3.2 1B} & \textbf{LLaMA 3.2 3B} \\ 
\midrule
\multirow{4}{*}{SVAMP} 
    & Original CoT & 5.56\textsubscript{±2.41}${\color{white}\uparrow}$ & \textbf{10.67}\textsubscript{±1.00}${\color{white}\uparrow}$ & \textbf{16.56}\textsubscript{±0.51}${\color{white}\uparrow}$ & 22.22\textsubscript{±0.69}${\color{white}\uparrow}$ & \textbf{67.11}\textsubscript{±1.35}${\color{white}\uparrow}$ & 52.44\textsubscript{±1.71}${\color{white}\uparrow}$ & 69.78\textsubscript{±1.07}${\color{white}\uparrow}$ \\ 
    & Least-to-most & \textbf{6.11}\textsubscript{±1.07}${\color{red}\uparrow}$ & 10.44\textsubscript{±0.69}${\color{teal}\downarrow}$ & 14.67\textsubscript{±1.00}${\color{teal}\downarrow}$ & 24.00\textsubscript{±1.45}${\color{red}\uparrow}$ & 66.56\textsubscript{±0.69}${\color{teal}\downarrow}$ & 54.44\textsubscript{±1.26}${\color{red}\uparrow}$ & \textbf{75.00}\textsubscript{±0.67}${\color{red}\uparrow}$ \\ 
    & RaR & 4.89\textsubscript{±0.19}${\color{teal}\downarrow}$ & 9.00\textsubscript{±0.58}${\color{teal}\downarrow}$ & 14.11\textsubscript{±0.69}${\color{teal}\downarrow}$ & \textbf{24.22}\textsubscript{±1.58}${\color{red}\uparrow}$ & 65.67\textsubscript{±1.73}${\color{teal}\downarrow}$ & \textbf{54.56}\textsubscript{±0.69}${\color{red}\uparrow}$ & 73.89\textsubscript{±1.26}${\color{red}\uparrow}$ \\ 
    & Symbolic CoT & 5.89\textsubscript{±0.51}${\color{red}\uparrow}$ & 6.44\textsubscript{±0.38}${\color{teal}\downarrow}$ & 9.00\textsubscript{±0.67}${\color{teal}\downarrow}$ & 19.22\textsubscript{±1.07}${\color{teal}\downarrow}$ & 64.78\textsubscript{±0.77}${\color{teal}\downarrow}$ & 51.78\textsubscript{±1.07}${\color{teal}\downarrow}$ & 72.89\textsubscript{±1.50}${\color{red}\uparrow}$ \\ 
\midrule
\multirow{4}{*}{GSM8K} 
    & Original CoT & \textbf{8.19}\textsubscript{±0.27}${\color{white}\uparrow}$ & 13.09\textsubscript{±0.83}${\color{white}\uparrow}$ & \textbf{16.86}\textsubscript{±1.25}${\color{white}\uparrow}$ & \textbf{22.47}\textsubscript{±0.86}${\color{white}\uparrow}$ & \textbf{53.42}\textsubscript{±0.83}${\color{white}\uparrow}$ & \textbf{39.58}\textsubscript{±1.04}${\color{white}\uparrow}$ & 62.29\textsubscript{±1.22}${\color{white}\uparrow}$ \\ 
    & Least-to-most & 7.88\textsubscript{±0.35}${\color{teal}\downarrow}$ & \textbf{13.52}\textsubscript{±0.87}${\color{red}\uparrow}$ & 15.54\textsubscript{±0.72}${\color{teal}\downarrow}$ & 21.86\textsubscript{±0.56}${\color{teal}\downarrow}$ & 51.93\textsubscript{±0.07}${\color{teal}\downarrow}$ & 39.25\textsubscript{±1.10}${\color{teal}\downarrow}$ & 62.07\textsubscript{±0.70}${\color{teal}\downarrow}$ \\ 
    & RaR & 5.89\textsubscript{±0.22}${\color{teal}\downarrow}$ & 10.84\textsubscript{±0.40}${\color{teal}\downarrow}$ & 13.72\textsubscript{±0.59}${\color{teal}\downarrow}$ & 20.02\textsubscript{±0.27}${\color{teal}\downarrow}$ & 51.99\textsubscript{±1.22}${\color{teal}\downarrow}$ & 38.09\textsubscript{±0.46}${\color{teal}\downarrow}$ & \textbf{63.02}\textsubscript{±0.56}${\color{red}\uparrow}$ \\ 
    & Symbolic CoT & 5.94\textsubscript{±0.62}${\color{teal}\downarrow}$ & 10.74\textsubscript{±0.64}${\color{teal}\downarrow}$ & 13.27\textsubscript{±0.20}${\color{teal}\downarrow}$ & 19.33\textsubscript{±0.73}${\color{teal}\downarrow}$ & 47.12\textsubscript{±0.39}${\color{teal}\downarrow}$ & 34.70\textsubscript{±0.89}${\color{teal}\downarrow}$ & 58.94\textsubscript{±0.83}${\color{teal}\downarrow}$ \\ 
    \midrule
\multirow{4}{*}{AQuA} 
    & Original CoT & 18.64\textsubscript{±1.98}${\color{white}\uparrow}$ & 21.92\textsubscript{±3.66}${\color{white}\uparrow}$ & 22.31\textsubscript{±1.38}${\color{white}\uparrow}$ & \textbf{25.85}\textsubscript{±0.45}${\color{white}\uparrow}$ & \textbf{44.88}\textsubscript{±0.79}${\color{white}\uparrow}$ & \textbf{33.20}\textsubscript{±2.17}${\color{white}\uparrow}$ & \textbf{47.24}\textsubscript{±4.77}${\color{white}\uparrow}$ \\ 
    & Least-to-most & 19.69\textsubscript{±3.22}${\color{red}\uparrow}$ & 20.73\textsubscript{±0.91}${\color{teal}\downarrow}$ & 23.10\textsubscript{±2.17}${\color{red}\uparrow}$ & 24.41\textsubscript{±1.42}${\color{teal}\downarrow}$ & 38.32\textsubscript{±1.86}${\color{teal}\downarrow}$ & 28.48\textsubscript{±2.17}${\color{teal}\downarrow}$ & 41.60\textsubscript{±2.77}${\color{teal}\downarrow}$ \\ 
    & RaR & \textbf{21.26}\textsubscript{±3.94}${\color{red}\uparrow}$ & \textbf{22.57}\textsubscript{±2.79}${\color{red}\uparrow}$ & \textbf{24.28}\textsubscript{±2.50}${\color{red}\uparrow}$ & 25.07\textsubscript{±3.94}${\color{teal}\downarrow}$ & 41.86\textsubscript{±3.16}${\color{teal}\downarrow}$ & 31.10\textsubscript{±2.39}${\color{teal}\downarrow}$ & 45.93\textsubscript{±4.60}${\color{teal}\downarrow}$ \\ 
    & Symbolic CoT & 16.14\textsubscript{±1.97}${\color{teal}\downarrow}$ & 19.16\textsubscript{±2.41}${\color{teal}\downarrow}$ & 21.00\textsubscript{±1.38}${\color{teal}\downarrow}$ & 20.87\textsubscript{±2.36}${\color{teal}\downarrow}$ & 40.94\textsubscript{±0.00}${\color{teal}\downarrow}$ & 28.08\textsubscript{±3.06}${\color{teal}\downarrow}$ & 42.13\textsubscript{±1.80}${\color{teal}\downarrow}$ \\ 
\midrule
\multirow{4}{*}{OBQA} 
    & Original CoT & 36.73\textsubscript{±0.76}${\color{white}\uparrow}$ & 46.07\textsubscript{±2.23}${\color{white}\uparrow}$ & 48.00\textsubscript{±1.40}${\color{white}\uparrow}$ & 54.87\textsubscript{±2.02}${\color{white}\uparrow}$ & 69.93\textsubscript{±1.36}${\color{white}\uparrow}$ & 63.60\textsubscript{±2.12}${\color{white}\uparrow}$ & 78.80\textsubscript{±0.80}${\color{white}\uparrow}$ \\ 
    & Least-to-most & 31.40\textsubscript{±1.74}${\color{teal}\downarrow}$ & 43.33\textsubscript{±2.02}${\color{teal}\downarrow}$ & 45.53\textsubscript{±2.39}${\color{teal}\downarrow}$ & 53.20\textsubscript{±2.50}${\color{teal}\downarrow}$ & 68.27\textsubscript{±1.03}${\color{teal}\downarrow}$ & 62.80\textsubscript{±2.23}${\color{teal}\downarrow}$ & 78.33\textsubscript{±1.62}${\color{teal}\downarrow}$ \\ 
    & RaR & \textbf{40.47}\textsubscript{±1.68}${\color{red}\uparrow}$ & \textbf{47.47}\textsubscript{±2.23}${\color{red}\uparrow}$ & \textbf{49.87}\textsubscript{±1.42}${\color{red}\uparrow}$ & \textbf{56.40}\textsubscript{±1.97}${\color{red}\uparrow}$ & \textbf{72.73}\textsubscript{±2.19}${\color{red}\uparrow}$ & \textbf{64.40}\textsubscript{±2.75}${\color{red}\uparrow}$ & \textbf{82.00}\textsubscript{±0.20}${\color{red}\uparrow}$ \\ 
    & Symbolic CoT & 31.67\textsubscript{±1.36}${\color{teal}\downarrow}$ & 35.13\textsubscript{±0.23}${\color{teal}\downarrow}$ & 37.73\textsubscript{±3.25}${\color{teal}\downarrow}$ & 41.13\textsubscript{±1.81}${\color{teal}\downarrow}$ & 61.80\textsubscript{±0.92}${\color{teal}\downarrow}$ & 52.47\textsubscript{±0.70}${\color{teal}\downarrow}$ & 72.13\textsubscript{±0.64}${\color{teal}\downarrow}$ \\ 
\bottomrule
\bottomrule
\end{tabular}
}
\caption{Performance of student models with different CoT formats. For each dataset, the best performance is boldfaced, and arrows show that the performance is increased ($\uparrow$) or decreased ($\downarrow$) over original CoT.}
\label{tab:arrows_only}
\end{table*}

\paragraph{Correlation between Granularity and Student Models}
\label{sec:Correlation of Granularity}

\begin{figure}[!htb] 
    \centering
    \begin{subfigure}[b]{0.49\columnwidth} 
        \centering
        \includegraphics[width=\columnwidth]{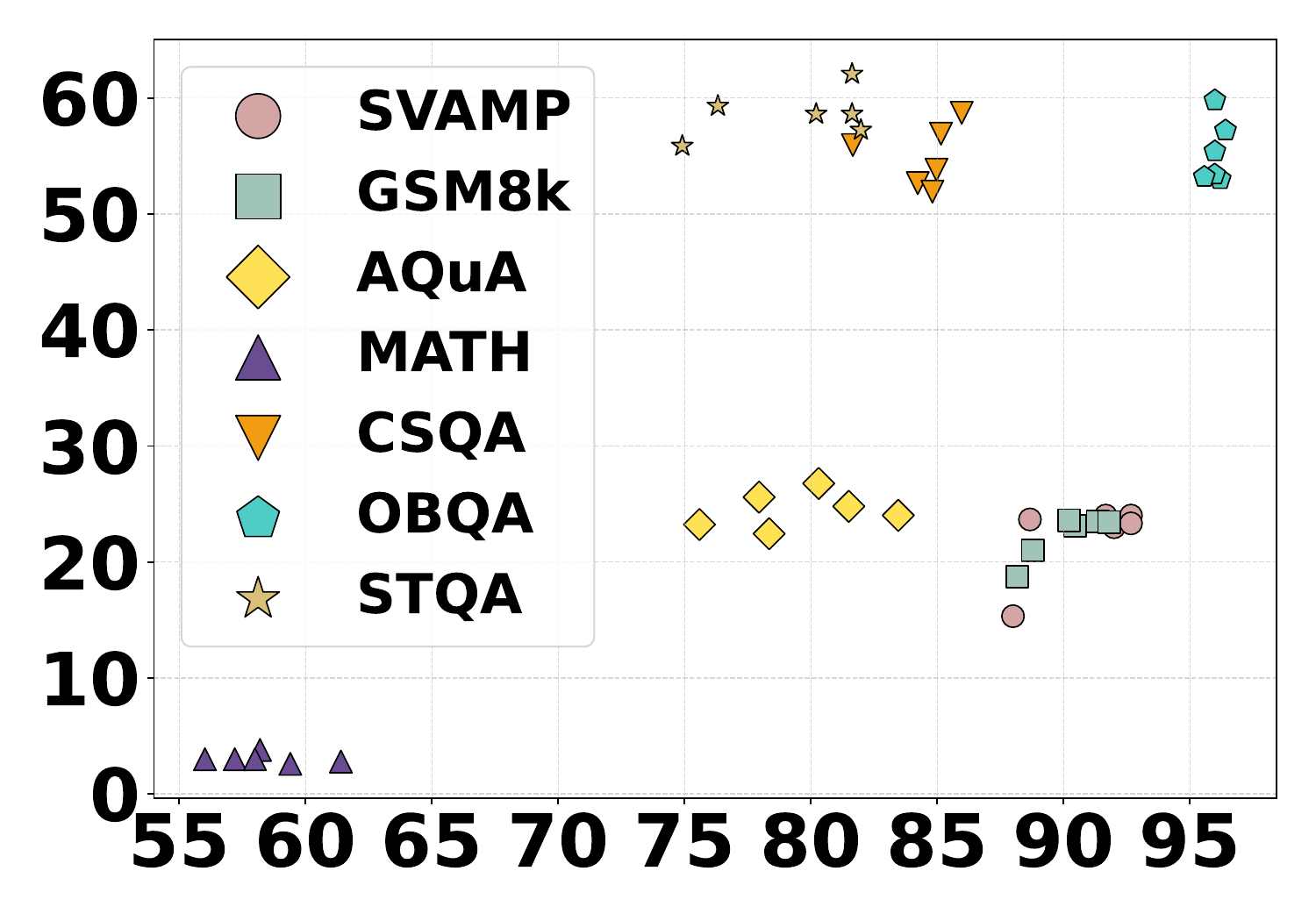}
        \caption{BLOOM 3B ($r=\textbf{0.52}$)}
    \end{subfigure}%
    \hfill
    \begin{subfigure}[b]{0.49\columnwidth} 
        \centering
        \includegraphics[width=\columnwidth]{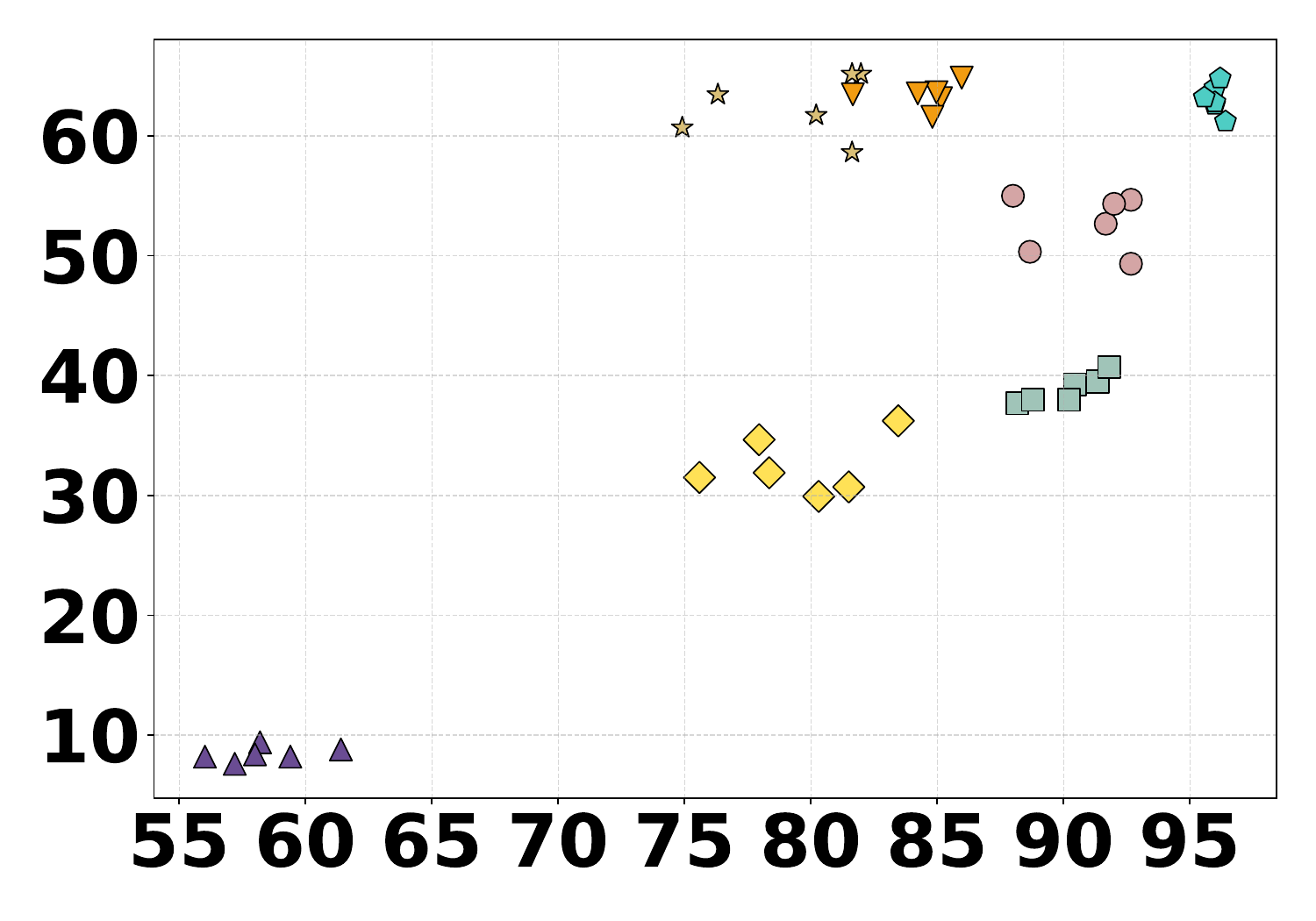}
        \caption{LLaMA 3.2 1B ($r=\textbf{0.75}$) }
    \end{subfigure}

    \begin{subfigure}[b]{0.49\columnwidth} 
        \centering
        \includegraphics[width=\columnwidth]{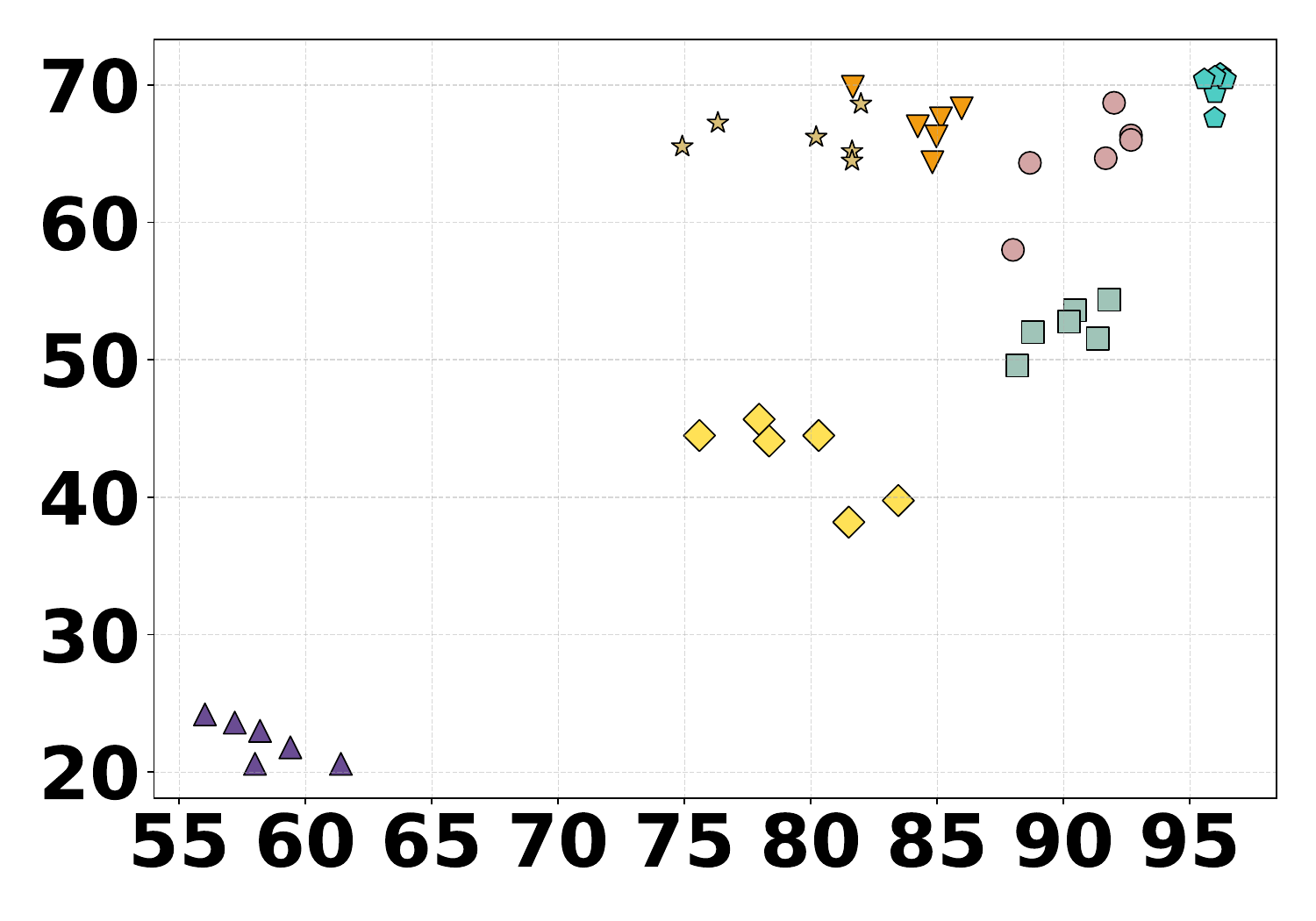}
        \caption{Gemma 2B ($r=\textbf{0.82}$)}
    \end{subfigure}%
    \hfill
    \begin{subfigure}[b]{0.49\columnwidth} 
        \centering
        \includegraphics[width=\columnwidth]{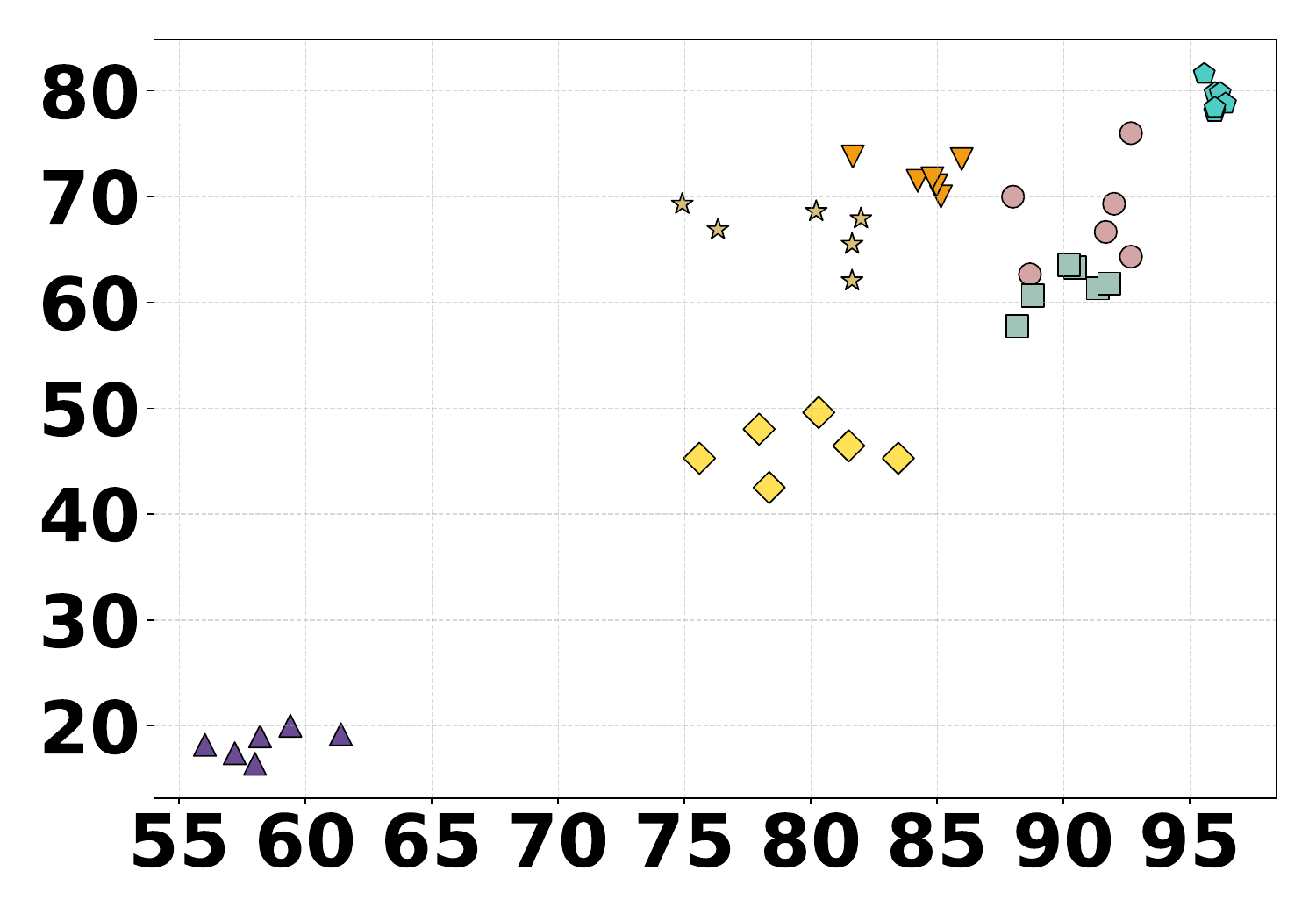}
        \caption{LLaMA 3.2 3B ($r=\textbf{0.89}$)}

    \end{subfigure}
    \caption{Scatter plots of teacher model (GPT-4o, x-axis) vs. student accuracy (y-axis) across datasets and granularity levels. Each point marker represents a specific dataset.}
    \label{fig:teacher_student_scatter}
\end{figure}

Figure~\ref{fig:teacher_student_scatter} illustrates the overall relationship between teacher and student performance across datasets at varying granularity levels. We further list the Pearson correlation coefficient ($r$) for each student model. The results reveal a clear trend: \emph{as student model capacity increases, its performance aligns more closely with the teacher model's preferences for reasoning granularity} 

Stronger student models demonstrate significantly higher alignment with the teacher’s optimal granularity, indicating better transferability of reasoning structures. In contrast, weaker student models show a lower correlation, suggesting the limited ability to adapt to the teacher’s granularity preferences. This highlights the importance of tailoring granularity configurations to \emph{match the capabilities of student models}, rather than relying solely on the teacher’s performance trends.  

\mybox{{\bf Conclusion}}{gray!40}{gray!10}{Increasing CoT granularity does not lead to monotonic improvements in student models. Stronger models benefit from higher granularity, whereas weaker models peak at intermediate levels and struggle with complex tasks. Optimizing granularity based on student capacity, rather than uniformly following the teacher model, is key to maximizing CoT distillation efficiency. }

\section{Effects of Format}
\label{sec:format}

Beyond granularity, the format of reasoning has been believed to influence model performance in prior research. However, SLMs often face limitations in processing complex reasoning structures. This raises a research question: \emph{Do these alternative formats improve student model performance, or are their benefits task-specific and limited?}

\paragraph{Choice of Reasoning Formats} In this section, we systematically evaluate the impact of alternative reasoning structures on student model performance. Since student models tend to perform stably at intermediate granularity levels, we let GPT-4o modify the format of the original CoT without changing the granularity (More details can be seen in Appendix~\ref{appendix:Format}). We compare the original CoT format with three alternative structures:

\begin{itemize}
\item \textbf{Least-to-most} (\citealt{zhou2023leasttomost}): A reasoning approach that decomposes a complex problem into a sequence of sub problems. Least-to-most excels in systematically breaking down problems into manageable parts to facilitate understanding and solution synthesis.
\item \textbf{Rephrase and Respond} (RaR)~\cite{deng2024rephraserespondletlarge}: A method where questions are rephrased to reduce ambiguity before answering, enabling iterative clarification and improving the LLM’s ability to respond accurately to nuanced queries.
\item \textbf{Symbolic CoT}~\cite{xu2024faithfullogicalreasoningsymbolic}: A reasoning structure that combines symbolic logic and CoT prompting, translating natural language into symbolic expressions for step-by-step logical deduction, enhancing faithfulness and flexibility in problem-solving.
\end{itemize}

Our results, summarized in Table~\ref{tab:arrows_only}, highlight a clear trend: \emph{the original CoT format often outperforms more complex or modified structures, primarily due to its simplicity and adaptability.} This contrasts with previous findings in LLMs, where these alternative formats frequently yield improvements. For SLMs, however, the added complexity of alternative formats generally increases cognitive load and hardly improve the performance.

\paragraph{Task-Specific Gains} While most tasks favor the original CoT format, certain alternative structures offer measurable benefits for specific scenarios. For example, RaR improves commonsense reasoning tasks by reducing ambiguity and enabling iterative clarification.
Least-to-most sometimes excels in mathematical reasoning by breaking problems into logical steps or symbolic expressions. \footnote{A possible reason for the suboptimal performance of Symbolic CoT is analyzed with examples in the Appendix~\ref{appendix:Symbolic CoT}.}

\paragraph{Model-Specific Trends} Stronger student models, such as LLaMA 3.2 3B, show improved performance under alternative formats, leveraging structural cues to refine problem-solving processes. However, these improvements are tied to specific tasks and do not generalize across all datasets.

Overall, our findings indicate that while CoT formats can occasionally enhance performance, their benefits are often task-dependent and come at the cost of introducing additional tokens. Moreover, SLMs have relatively limited pretraining corpora, which likely contain fewer instances of these reasoning formats. As a result, weaker models struggle to effectively learn and utilize them, making it even harder for CoT format variations to yield consistent improvements.
Given these observations, we argue that \emph{adjusting CoT formats alone may not be the most effective approach for improving SLM performance}. This contrasts with the consistent impact of granularity, as highlighted in Section~\ref{sec:granularity}, suggesting that \emph{focusing on granularity is a more effective strategy than altering CoT formats}. 

\mybox{{\bf Conclusion}}{gray!40}{gray!10}{While alternative CoT formats sometimes offer some task-specific benefits, the original CoT format from teacher models often remains the most effective for general-purpose SLM training.}

\section{Effects of Teacher Model}
\label{sec:teacher}

In CoT distillation, teacher models serve as the source of CoT reasoning annotations for training student models. Prior research of KD assumes that teacher models with better performance naturally lead to better student models ~\cite{zong2023better}. This assumption stems from the belief that higher-performing teachers generate more accurate reasoning steps and answers, which, when distilled into student models, enhance their capabilities. However, this assumption may not hold universally as SLMs might have limited capacity to replicate the reasoning complexity of strong teachers. 

In this section, we analyze the performance of student models under four teachers, GPT-4o, LLaMA 3 70B, Gemini-1.5-Flash and the human expert. 
We aim to determine \emph{whether the choice of teacher affects the ability of student models to effectively distill and replicate CoT reasoning}. 

\begin{figure}[!htb] 
    \centering
    \begin{subfigure}[b]{0.49\columnwidth} 
        \centering
        \includegraphics[width=\columnwidth]{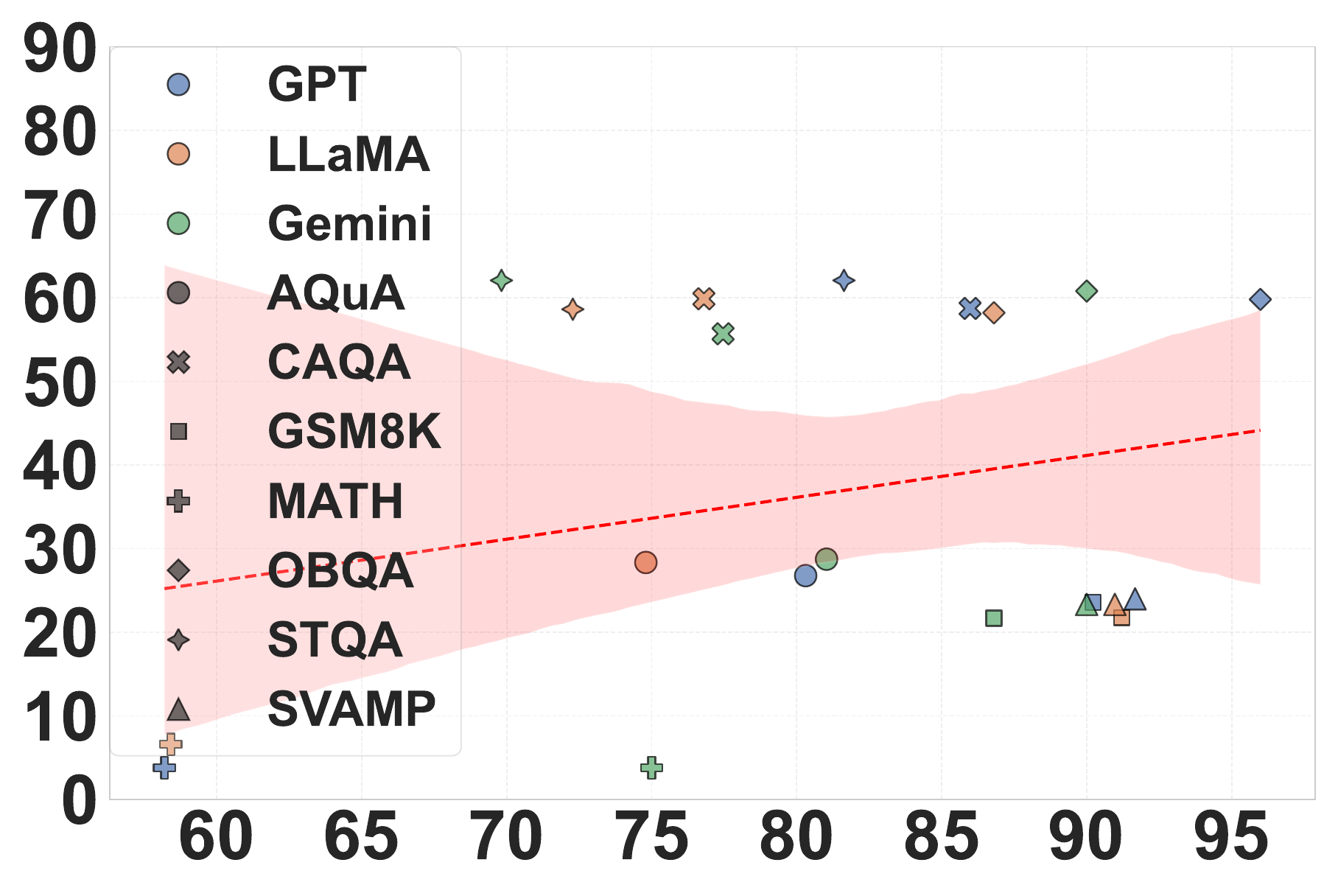}
        \caption{BLOOM 3B ($r=\textbf{0.25}$)}
    \end{subfigure}%
    \hfill
    \begin{subfigure}[b]{0.49\columnwidth} 
        \centering
        \includegraphics[width=\columnwidth]{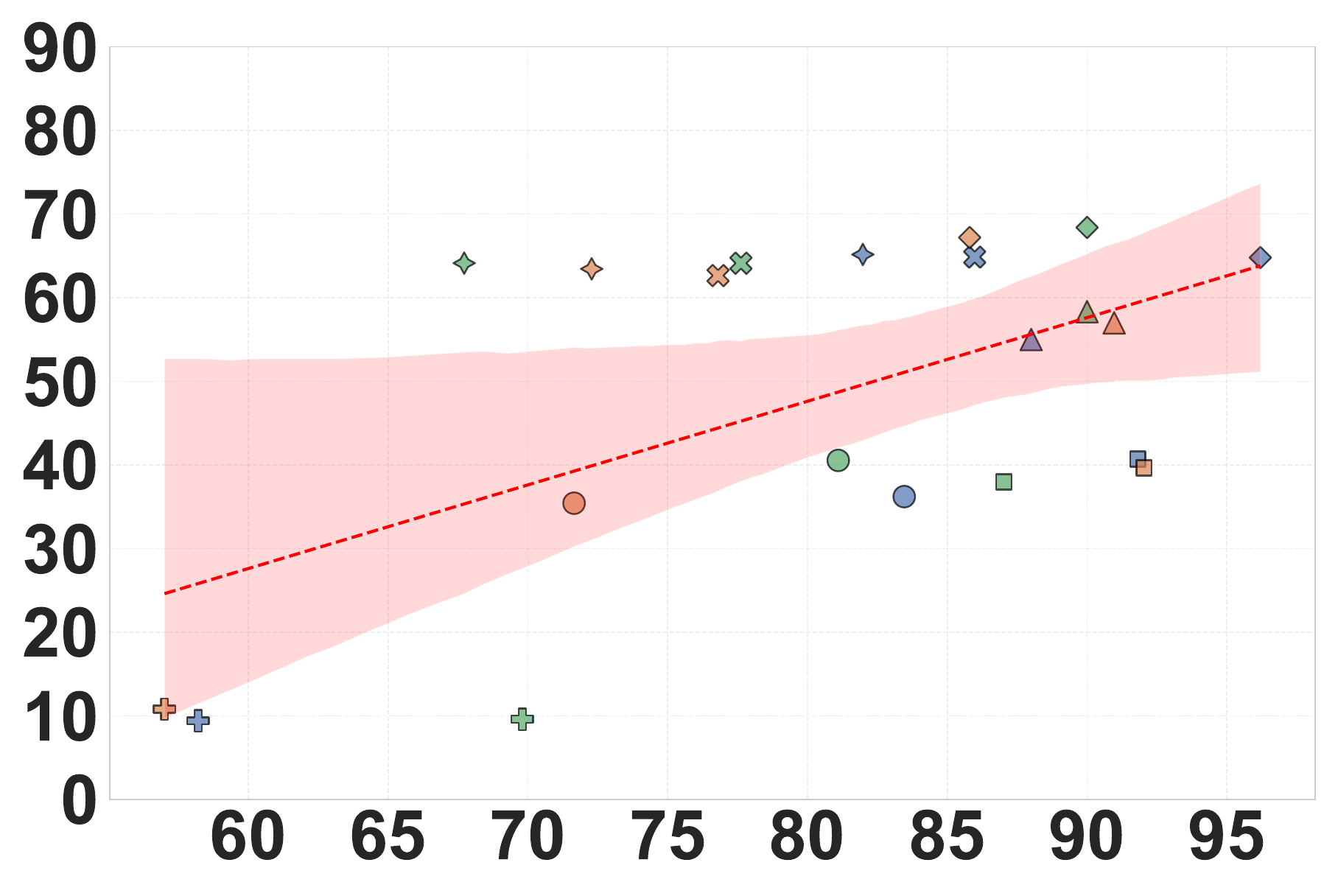}
        \caption{LLaMA 3.2 1B ($r=\textbf{0.56}$)}
    \end{subfigure}

    \begin{subfigure}[b]{0.49\columnwidth} 
        \centering
        \includegraphics[width=\columnwidth]{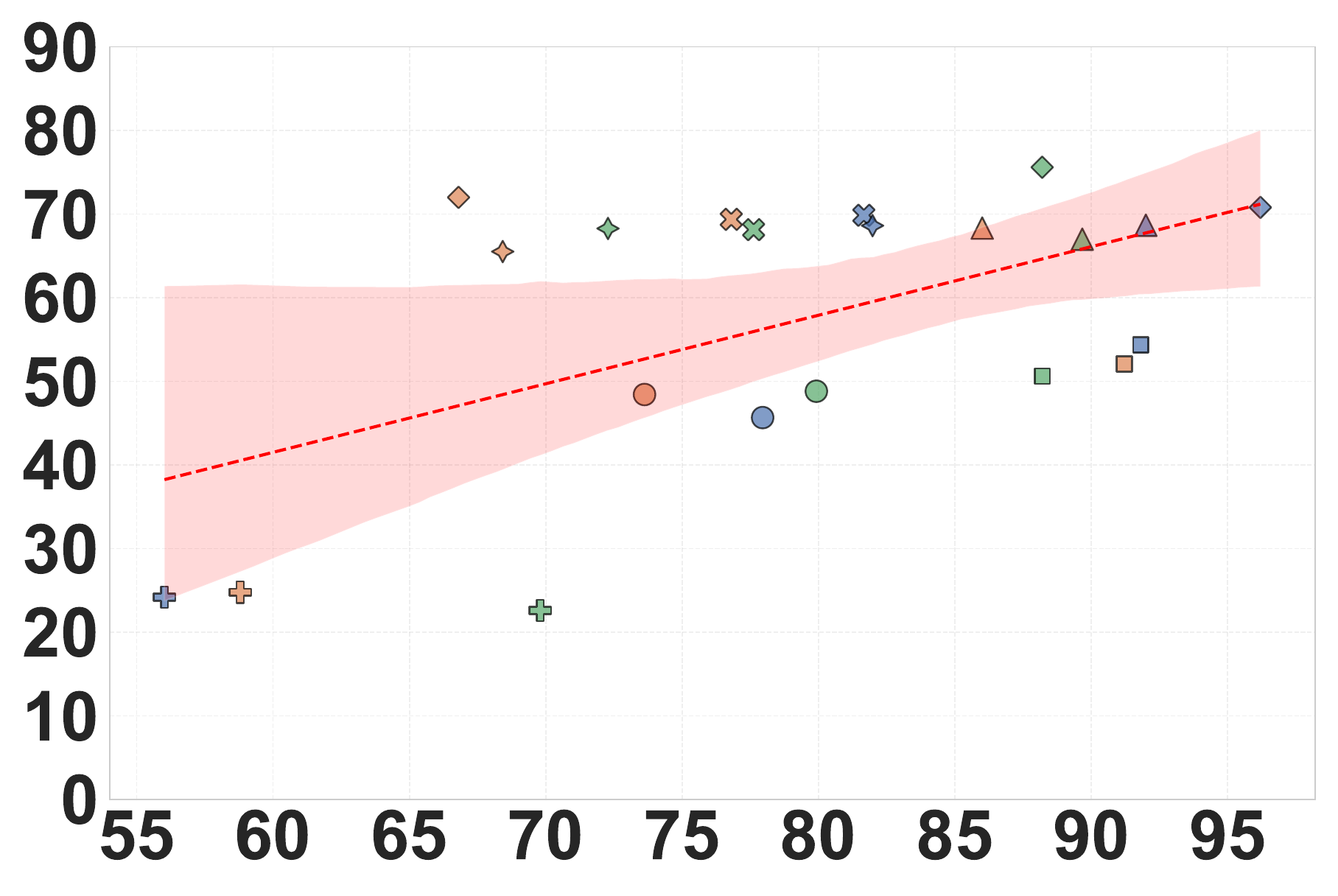}
        \caption{Gemma 2B ($r=\textbf{0.55}$)}
    \end{subfigure}
    \hfill
    \begin{subfigure}[b]{0.49\columnwidth} 
        \centering
        \includegraphics[width=\columnwidth]{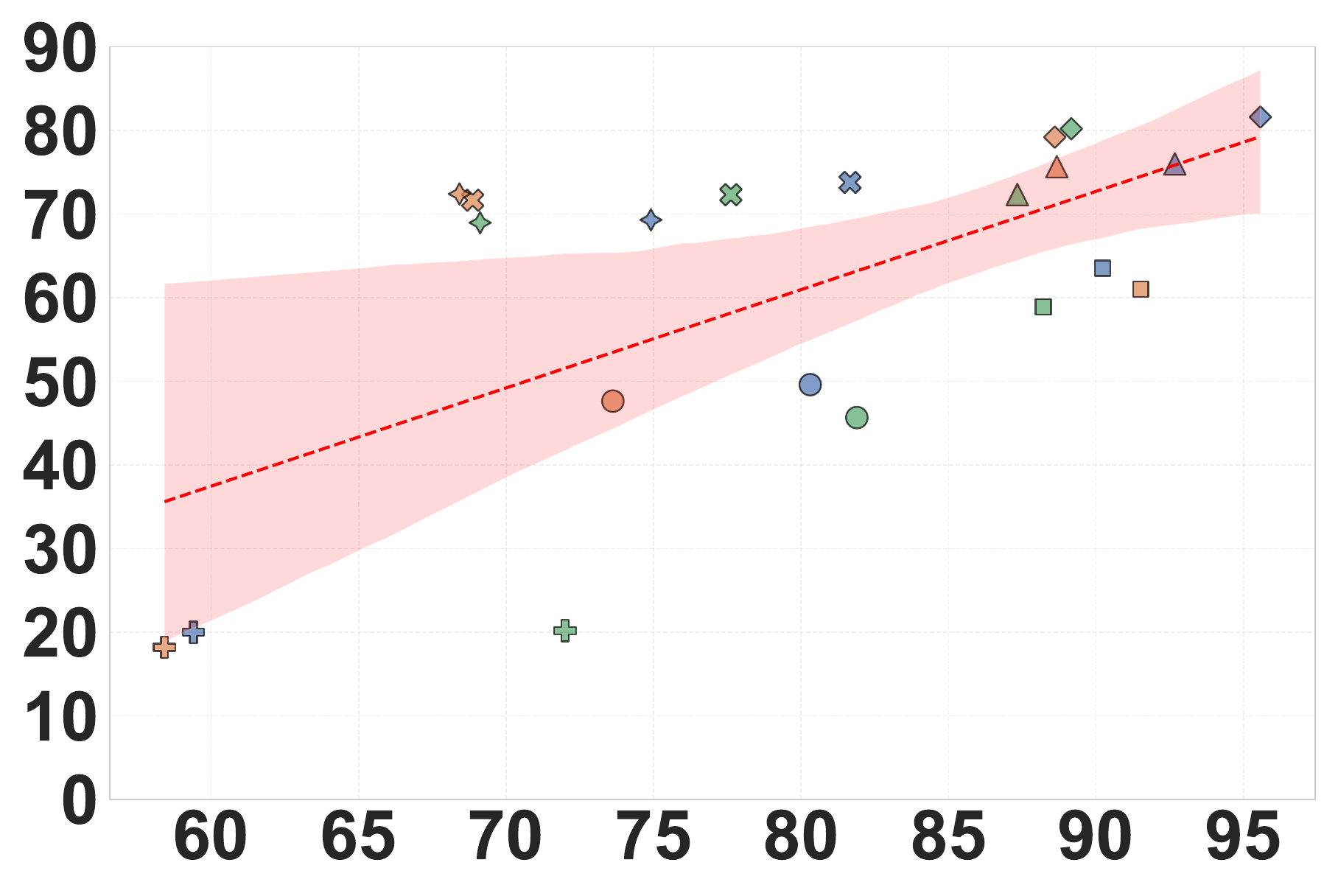}
        \caption{LLaMA 3.2 3B ($r=\textbf{0.64}$)}
    \end{subfigure}

    \caption{Scatter plots of teacher (x-axis) vs. student model accuracy (y-axis) across datasets. 
    GPT refers to GPT-4o, LLaMA refers to LLaMA 3 70B, and Gemini refers to Gemini-1.5-Flash. }
    \label{fig:scattergemma}
\end{figure}

\paragraph{Is Higher Teacher Accuracy Always Better?} We first investigate whether a higher teacher accuracy directly translates into improved student performance. As shown in Figure~\ref{fig:scattergemma}, we selected the best-performing student model for each dataset under different teacher models’ CoT supervision. It can be observed that points closer to the right side of the x-axis are not always positioned near the top of the y-axis. This indicates that, contrary to intuitive expectations, while a reasonably accurate teacher can effectively impart essential reasoning patterns, excessively high teacher accuracy does not always yield proportional improvements in student accuracy. \emph{Teacher accuracy alone is not the determining factor for student performance}, which aligns with our findings in Section~\ref{sec:Correlation of Granularity}. 

Moreover, we do not observe a significant preference pattern within the same model family. However, we find that \emph{stronger student models tend to benefit more significantly when trained under the guidance of stronger teacher models}. This suggests that the trade-off between teacher model capability and computational cost should be carefully adjusted based on the target student model’s capacity. For simpler tasks, a less advanced teacher model is often sufficient, producing results comparable to those obtained from more powerful, computationally expensive teachers.

\begin{figure}[!htb] 
    \centering
    \begin{subfigure}[b]{0.49\columnwidth} 
        \centering
        \includegraphics[width=\columnwidth]{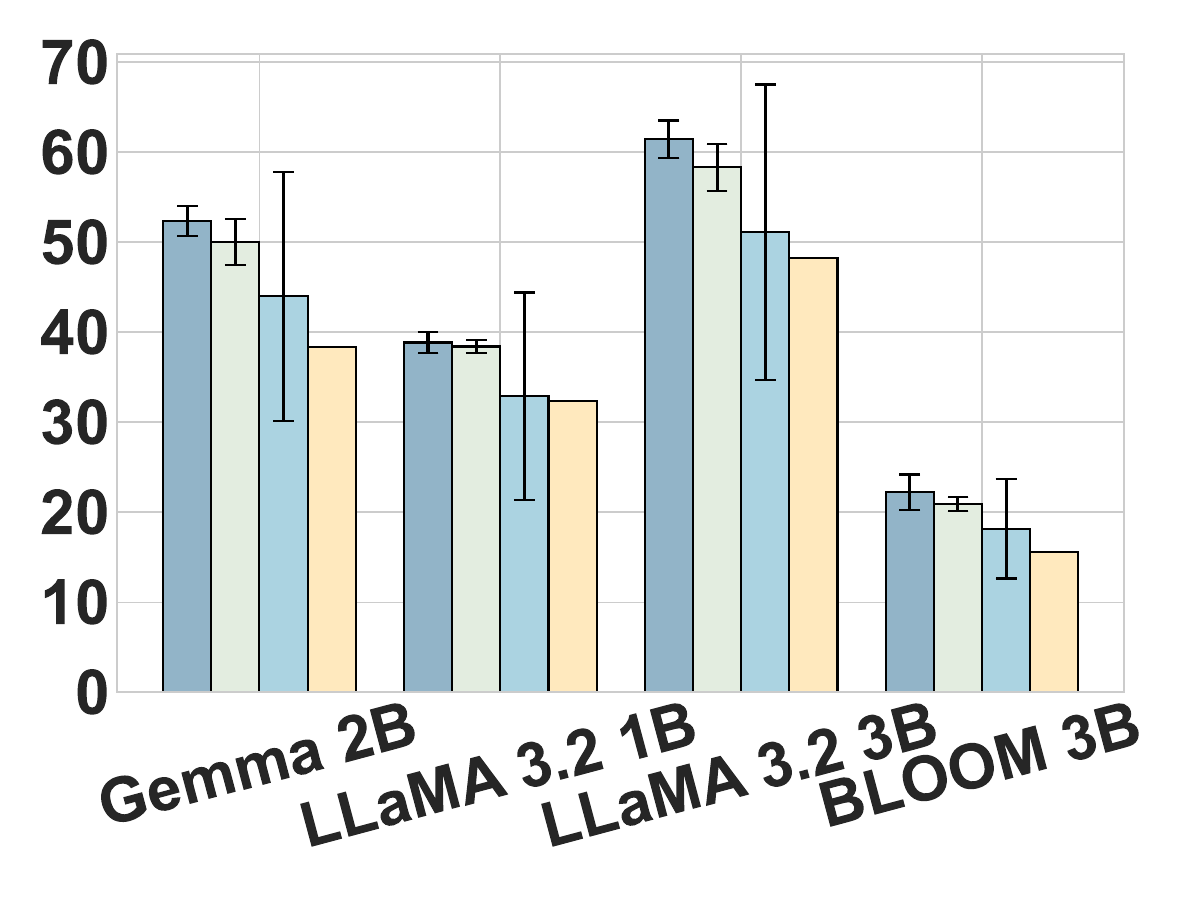}
        \caption{GSM8K}
    \end{subfigure}%
    \hfill
    \begin{subfigure}[b]{0.49\columnwidth} 
        \centering
        \includegraphics[width=\columnwidth]{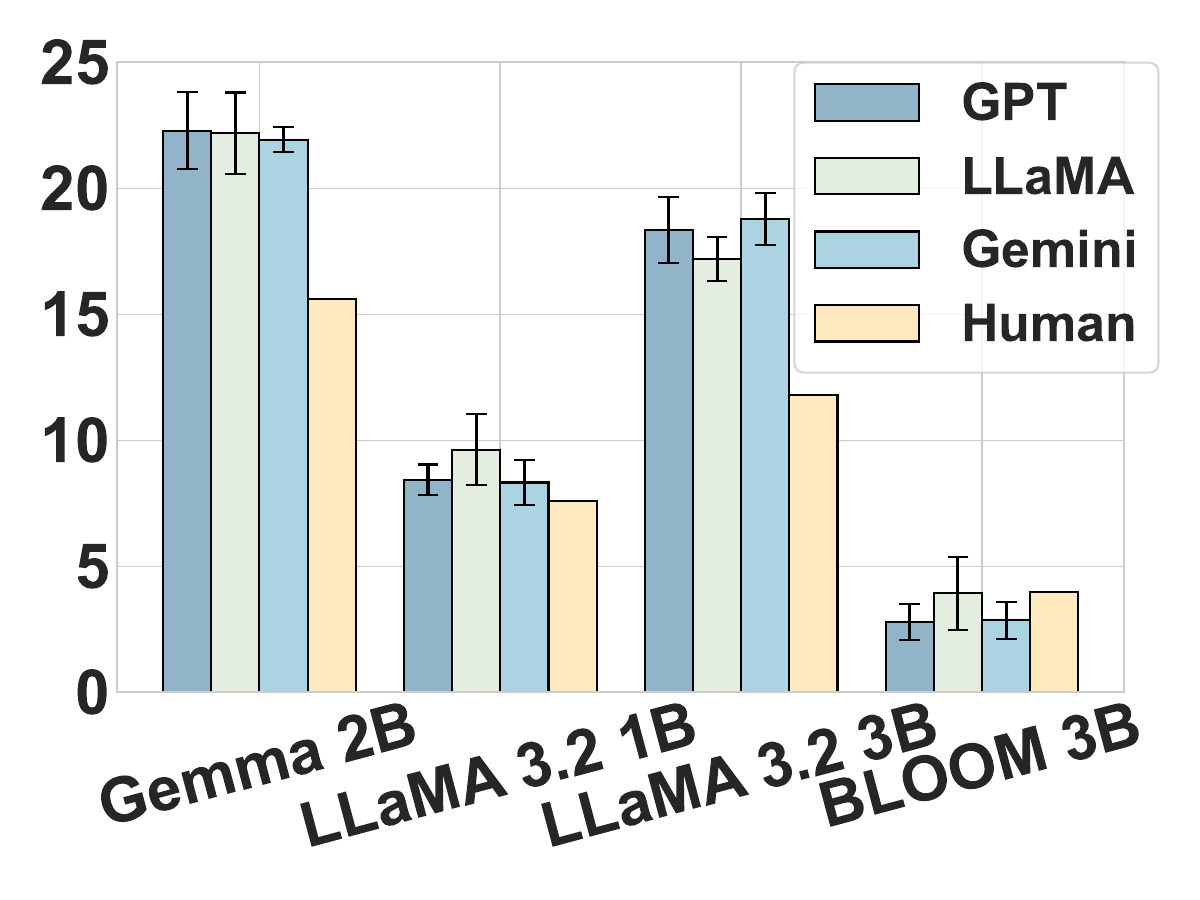}
        \caption{MATH}
    \end{subfigure}

    \begin{subfigure}[b]{0.49\columnwidth} 
        \centering
        \includegraphics[width=\columnwidth]{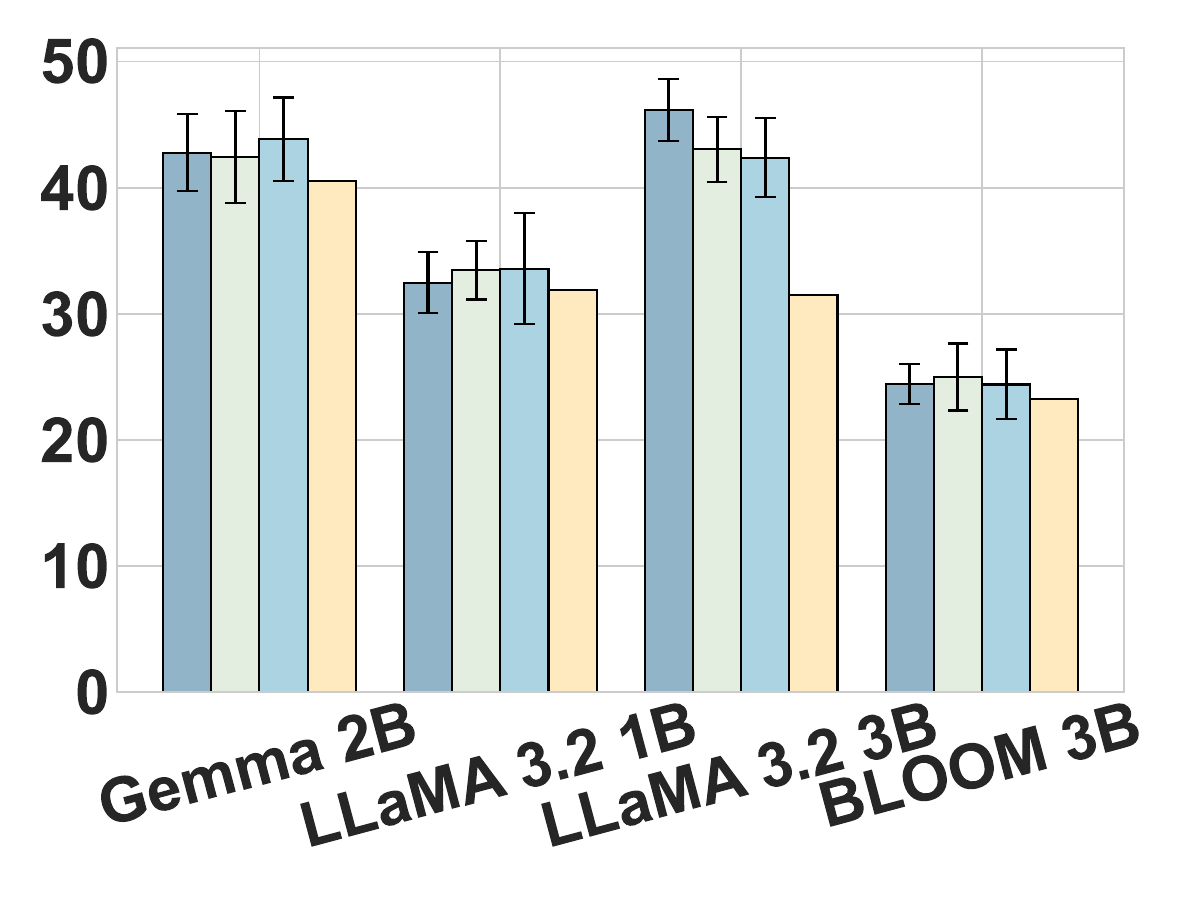}
        \caption{AQuA-RAT}
    \end{subfigure}%
    \hfill
    \begin{subfigure}[b]{0.49\columnwidth} 
        \centering
        \includegraphics[width=\columnwidth]{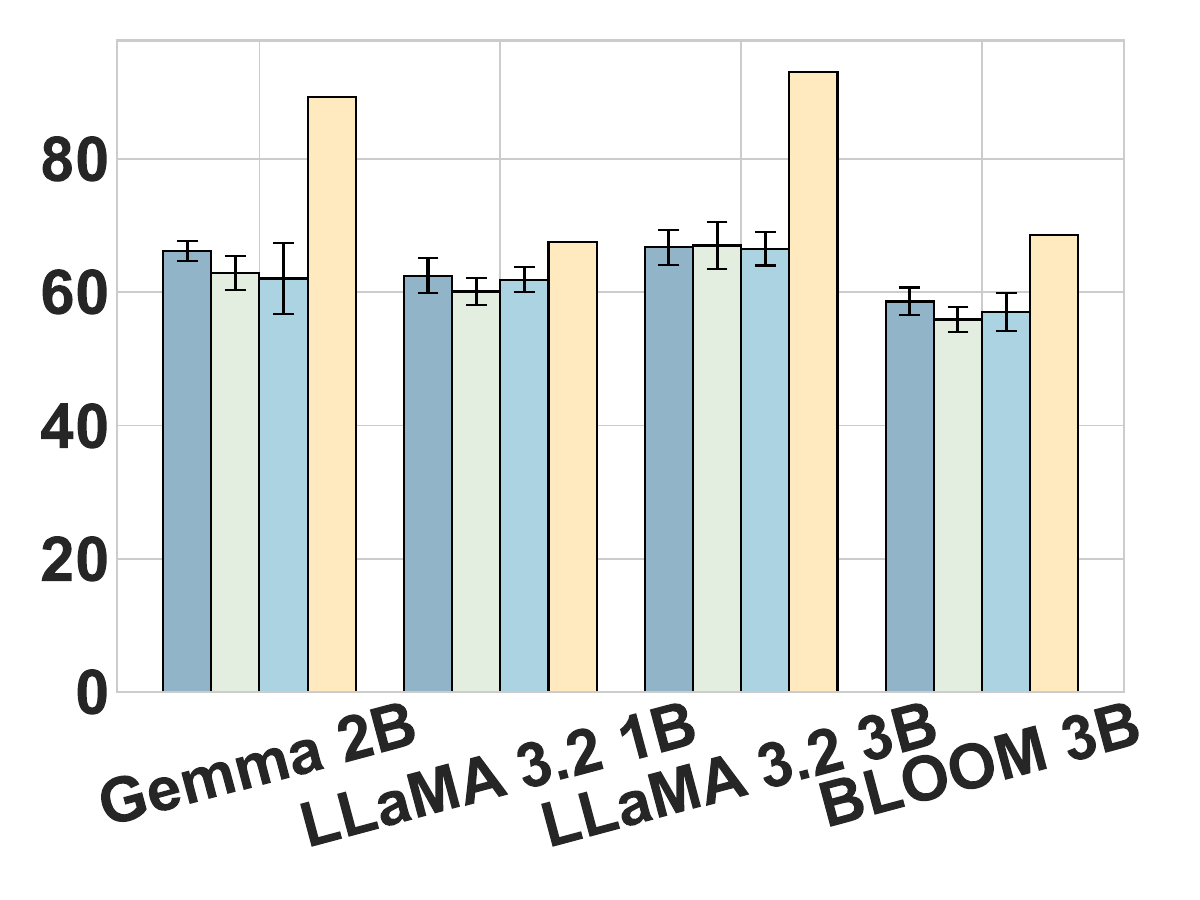}
        \caption{StrategyQA}
    \end{subfigure}

    \caption{Student model performance across different teacher models. Each bar represents the average accuracy of a specific student model trained on CoT from different teacher models. 
    }
    \label{fig:teacherstyle}
\end{figure}

\paragraph{Human vs LLM: Task-Specific Effectiveness}
As seen in Figure~\ref{fig:teacherstyle}, in mathematical reasoning tasks, student models achieve higher accuracy when fine-tuned on LLM-generated CoTs compared to human annotations, although the accuracy of the teacher model itself performs poorly on difficult mathematical datasets compared with human-labeled data. Conversely, for commonsense reasoning tasks like StrategyQA, human-annotated CoTs dramatically improve student model performance. This phenomenon arises because LLMs generate \emph{structured and detailed reasoning chains} that closely align with the symbolic and procedural nature of mathematical tasks. In contrast, human-annotated CoTs often lack the rigorous step-by-step structure required for effective mathematical reasoning. However, Human annotations excel at capturing \emph{nuanced contextual understanding, creative inferences, and interpretive reasoning}, which are crucial for tasks involving ambiguous or open-ended questions.
These findings underscore \emph{the importance of selecting CoT sources based on task characteristics} rather than assuming a universal superiority of either LLMs or human annotations. Full results can be found in Appendix~\ref{appendix:Teacheracc}. 

\begin{figure}[!htb] 
    \centering
    \begin{subfigure}[b]{0.49\columnwidth} 
        \centering
        \includegraphics[height=2.3cm]{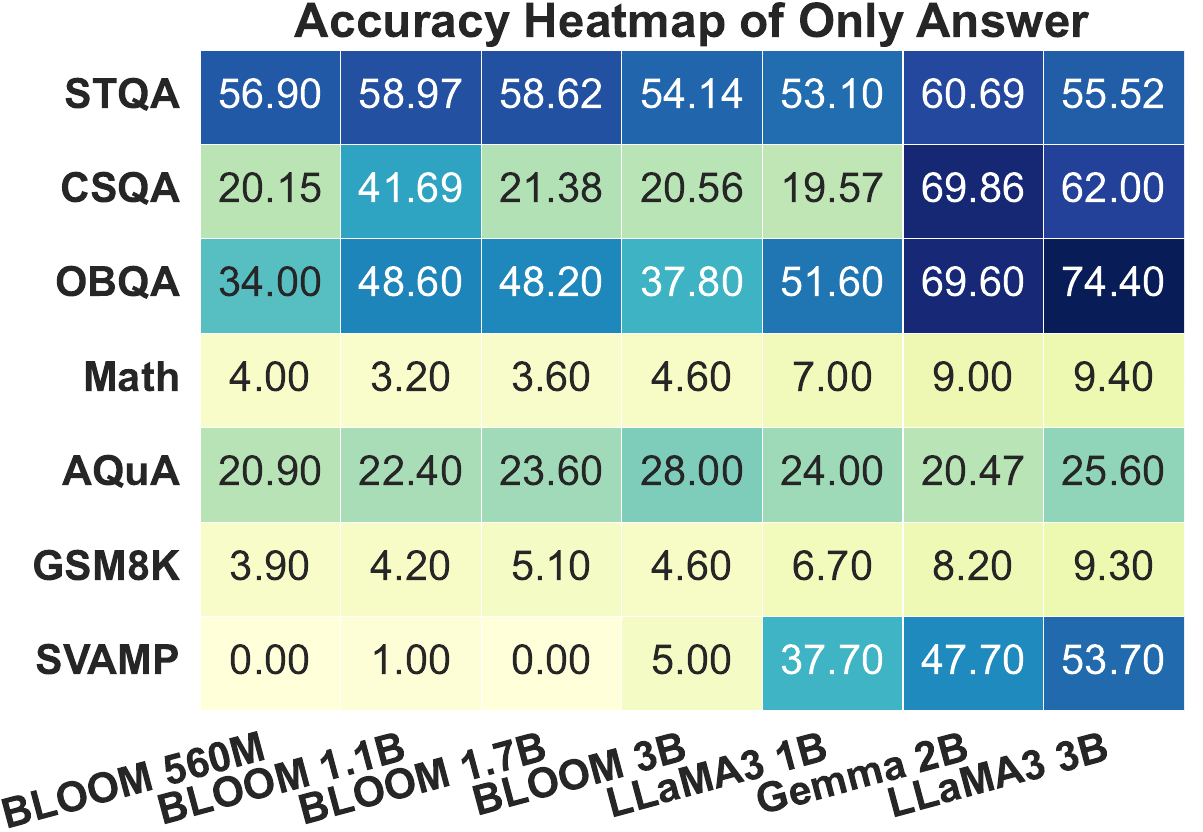}
    \end{subfigure}%
    \hfill
    \begin{subfigure}[b]{0.49\columnwidth} 
        \centering
        \includegraphics[height=2.3cm]{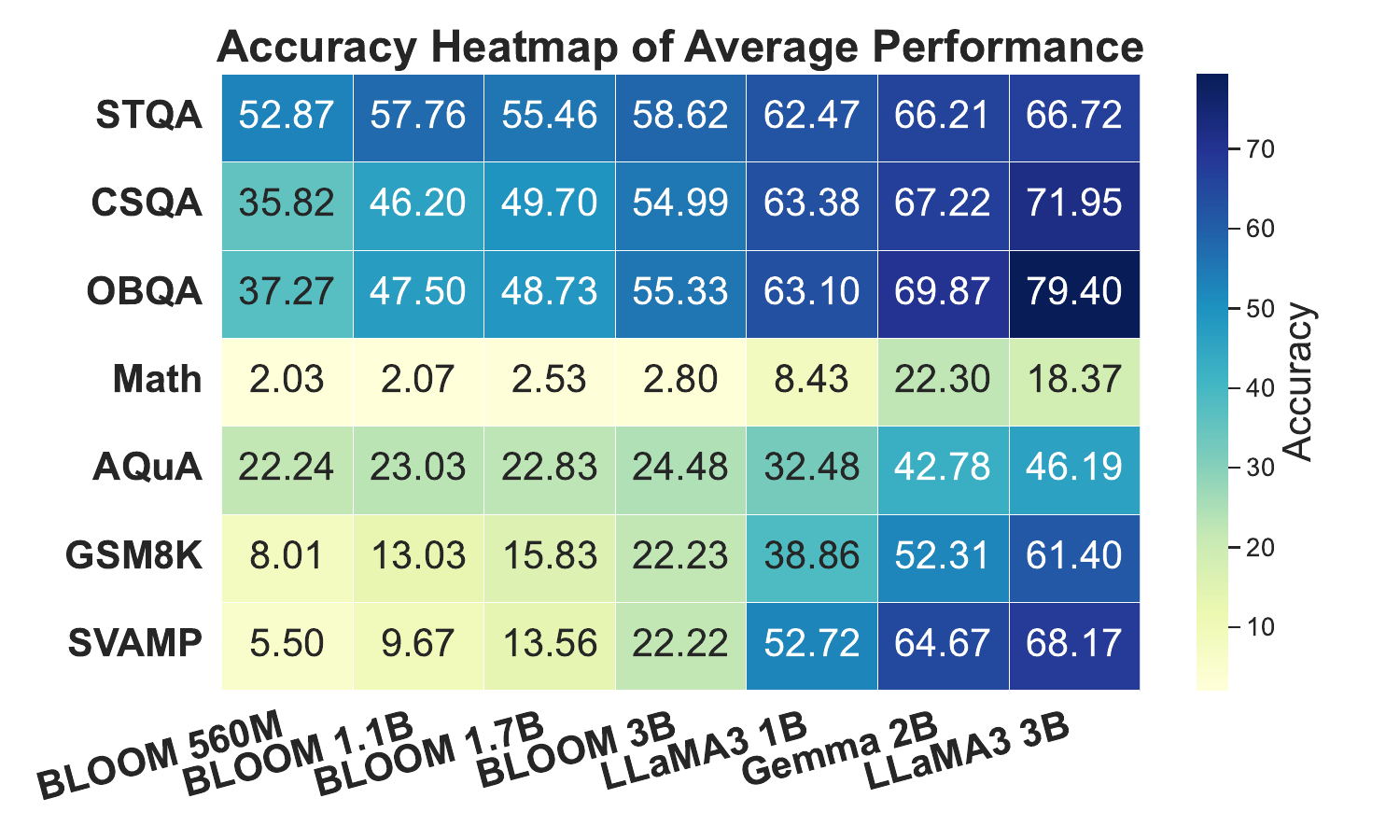}
    \end{subfigure}
    \caption{The \textit{Only Answer} heatmap represents the baseline accuracy of student models without reasoning supervision, while the \textit{Average Performance} heatmap shows the average accuracy of student models trained on CoT from ChatGPT-4o. }
    \label{fig:heatmap}
\end{figure}

\paragraph{The Matthew Effect in SLMs} 
We explore the relationship between student model capacity and the benefits gained from CoT distillation, shedding light on the uneven distribution of performance improvements across models of varying capabilities.
Figure~\ref{fig:heatmap} presents two heatmaps comparing student model performance before and after CoT distillation. The results reveal a \emph{Matthew Effect}: \emph{stronger student models achieve greater performance gains from CoT distillation than weaker models}, demonstrating their potential ability to leverage detailed reasoning steps. 
This phenomenon aligns with Vygotsky's Zone of Proximal Development (ZPD)~\cite{vygotsky1978mind}, where weaker student models have a narrower ZPD, limiting their ability to absorb complex CoT reasoning. If reasoning complexity is too high relative to a model's ZPD, it may fail to extract useful patterns, limiting the effectiveness of CoT distillation. In contrast, stronger models have a wider ZPD, enabling them to integrate and generalize from multi-step reasoning. CoT distillation provides gains on more challenging ones, where their capacity allows them to fully leverage structured reasoning. This highlights the need for adaptive CoT supervision, where reasoning depth is modulated based on the student's ability to process and learn from it.

\mybox{{\bf Conclusion}}{gray!40}{gray!10}{The assumption that a better teacher always produces a better student does not universally hold for SLMs. Stronger student models benefit more from advanced teacher models. Teacher choice should be task-specific: LLM-generated CoTs improve mathematical reasoning, while human annotations excel in commonsense reasoning.}

\section{Conclusion}
This study systematically examined key factors influencing CoT distillation in SLMs, including teacher selection, granularity, and format. First, We found that finer-grained CoT benefits stronger SLMs, and weaker models perform better with simpler annotations. Then,  while CoT format significantly impacts LLMs, its effect on SLMs is more subtle. Importantly, better teacher models do not always yield better students, as the effectiveness of CoT distillation depends on a model’s ability to absorb reasoning complexity within its ZPD. Notably, human-annotated CoTs underperform on mathematical tasks but can surpass LLM-generated CoTs in certain commonsense reasoning tasks. Overall, CoT distillation proves more effective for stronger SLMs and complex tasks, emphasizing the need for tailored granularity and teacher selection strategies to optimize reasoning performance in resource-constrained settings.

\section*{Limitations}
Despite the promising results of our study, several limitations must be acknowledged. First, during data generation and testing, some tasks triggered safety concerns in the models, causing them to refuse to generate CoTs. In these cases, we resorted to directly using the provided answers for fine-tuning, which may have constrained the diversity and quality of the reasoning chains, potentially affecting the distillation outcomes. Second, the ability of teacher models to generate CoTs is inherently tied to their reasoning capabilities. For certain tasks, teacher models were unable to reverse-engineer plausible CoTs from the given answers due to their limited capabilities, resulting in incomplete or suboptimal reasoning chains. Lastly, this study did not focus on exploring novel KD techniques but instead aimed to systematically analyze the effects of existing approaches on CoT granularity, format, and teacher selection. We recognize that alternative training strategies may influence distillation outcomes. These limitations highlight the need for further exploration of CoT generation methods and distillation techniques tailored to task complexity and student capabilities. Future work may benefit from integrating curriculum-based strategies or lightweight multi-stage distillation protocols to enhance the fidelity and generalizability of SLMs.

\section*{Ethics Statement}
The datasets used in our experiment are publicly
released and labeled through interaction with humans in English. In this process, user privacy is protected, and no personal information is contained in the dataset. The scientific artifacts that we used are available for research with permissive licenses. And the use of these artifacts in this paper is consistent with their intended use. Therefore, we believe that our research work meets the ethics of ACL.

\section*{Acknowledgement}
We thank EIT and IDT High Performance Computing Center for providing computational resources for this project. This work was supported by the Research Grants Council of Hong Kong (PolyU/15209724), Innovation Consortium Program for Green and Efficient Intelligent Appliance of Ningbo City (No.2022H002), Young Tech Innovation Leading Talent Program of Ningbo City (No.2023QL008) and  2035 Key Research and Development Program of Ningbo City under Grant No.2024Z127.

\bibliography{custom}

\clearpage 
\newpage
\appendix
\onecolumn

\section*{Appendix}
\section{Overview of Training and Test Datasets}  
\label{appendix:dataset}
For our experiments, we used three models (Llama3 70B, Gemini-1.5-Flash, GPT-4o) on multiple existing datasets, including mathematical reasoning datasets (SVAMP, GSM8K, AQuA-RAT, MATH) and commonsense reasoning datasets (OpenBookQA, CommonsenseQA, and StrategyQA) to generate CoT outputs. Table \ref{tab:Overview of Training and Test Datasets} shows the overview of the training and test datasets~\cite{yue2024mammoth}. Table \ref{tab:human-annotation} shows some examples of our datasets.

\begin{table*}[h!]
\centering
\makebox[\textwidth]{ 
\resizebox{\textwidth}{!}{
\begin{tabular}{c|cc|c|c}
\midrule
\multirow{2}{*}{\textbf{Training Dataset}} & \multicolumn{2}{c|}{\textbf{Samples}} & \multirow{2}{*}{\textbf{Fields}} & \multirow{2}{*}{\textbf{Human Annotation}} \\ \cline{2-3}
                          & Training & Testing  &                 &                 \\ \hline
SVAMP                     & 700             & 300             & Arithmetic problems & Yes \\
\rowcolor{gray!30} GSM8K  & 7.4k            & 1.3k            & Grade-school math   & Yes \\
AQuA-RAT                  & 6.1k            & 254             & Algebraic reasoning, multi-step & Yes \\
\rowcolor{gray!30} Math   & 1.3k            & 500             & Pre-Algebra, Algebra, Counting \& Probability, Number Theory & Yes \\ 
CommonsenseQA             & 9.7k            & 1.2k            & Commonsense knowledge & Yes \\
\rowcolor{gray!30} OpenBookQA & 4.9k        & 500             & Domain-specific knowledge & No \\
StrategyQA                & 2k              & 290             & Multi-step reasoning & Yes \\
\midrule
\end{tabular}
} 
} 
\caption{Overview of Training and Test Datasets.}
\label{tab:Overview of Training and Test Datasets} 
\end{table*}

\definecolor{LightGray}{gray}{0.95}
\definecolor{LightCyan}{rgb}{0.88,1,1}

\begin{table*}[h!]
\centering
\setlength{\tabcolsep}{10pt} 
\renewcommand{\arraystretch}{1.5} 
\fontsize{10}{12}\selectfont 
\resizebox{\textwidth}{!}{
\begin{tabular}{p{0.15\linewidth} p{0.45\linewidth} p{0.45\linewidth}} 
\toprule
\rowcolor{LightGray}
\textbf{Dataset} & \textbf{Problem} & \textbf{Characteristics} \\ 
\midrule

\rowcolor{white}
SVAMP &  
There are 87 oranges and 290 bananas in Philip's collection.  
If the bananas are organized into 2 groups and oranges are  
organized into 93 groups How big is each group of bananas? &  
290.0 / 2.0 = 145.0.  
The answer is 145.0. \\

\rowcolor{gray!30}
GSM8K &  
Natalia sold clips to 48 of her friends in April, and then she  
sold half as many clips in May. How many clips did Natalia sell  
altogether in April and May? &  
Natalia sold 48/2 = <<48/2=24>>24 clips in May.  
Natalia sold 48+24 = <<48+24=72>>72 clips altogether  
in April and May. 72 \\ 

\rowcolor{white}
AQuA-RAT &  
A man can swim in still water at 7.5 km/h, but takes twice as  
long to swim upstream than downstream. The speed of the stream is? Answer Choices: (A) 3 (B) 2.5 (C) 2.25 (D) 1.5 (E) 4&  
M = 7.5 S = x DS = 7.5 + x  
US = 7.5 + x 7.5 + x = (7.5 - x)2  
7.5 + x = 15-2x 3x = 7.5  
x = 2.5 Answer: C \\

\rowcolor{gray!30}
Math &  
Find the sum of all positive divisors of 50 that are also  
divisors of 15. &  
The positive factors of 50 are 1, 2,5, 10, 25, 50.  
Of these, only 1 and 5 divide 15.  
Their sum is 1+5 = 6. \\

\rowcolor{white}
CommonsenseQA &  
Bill did not abandon the fight, but did what to the enemy?  
Answer choices: A: arrogate, B: retain, C: embrace,  
D: smile, E: engage &  
Bill engaged in a fight with enemy.  
Other options are not a type of fights  
one takes with enemy.  
The answer is E. \\

\rowcolor{gray!30}
StrategyQA &  
Are more people today related to Genghis Khan than Julius Caesar? &  
Julius Caesar had three children.  
Genghis Khan had sixteen children.  
Modern geneticists have determined that  
out of every 200 men today has DNA  
that can be traced to Genghis Khan.  
The answer is True. \\

\bottomrule
\end{tabular}
}
\caption{Examples of Human Annotation for All Datasets.}
\label{tab:human-annotation}
\end{table*}

\clearpage
\section{Training setup} 
\label{appendix:setup}
Our experiment uses the LLaMA-Factory framework \cite{zheng2024llamafactory} to fine-tune models, and the training parameters are as follows:

\begin{table}[h]
\centering
\small
\begin{tabular}{lc}
\toprule
\textbf{Parameter} & \textbf{Value} \\ 
\midrule
Learning Rate      & 3e-5        \\ 
Num Train Epochs   & 3            \\ 
LR Scheduler       & Cosine         \\ 
Max Grad Norm      & 1.0            \\ 
Optimizer          & AdamW          \\ 
Template           & gemma/alpaca/llama3 \\
\bottomrule
\end{tabular}
\caption{Configuration for training parameters.}
\label{tab:training_parameters}
\end{table}

\section{Different CoT Granularity Dataset Collection} 
\label{appendix:Granularity}
\subsection{Workflow} 
\label{appendix:Granularity_workflow}
\textbf{The granularity dataset processing steps are detailed below:}

\textbf{1. 0-Shot Example Generation:}
The same question is first provided to the three models using a 0-shot prompt. The models generate a single 1-shot example (including both a question and a corresponding output) as the output. This step  ensures that the models first generate a baseline example. The generated example serves as a guide for subsequent responses.

\textbf{2. Input Construction:}
Each question is provided to the models, along with its corresponding ground-truth answer from the original dataset and the generated 1-shot example (from Step 1). Including the 1-shot example in the input establishes a reference point for the model, enhancing coherence and quality of generated outputs.

\textbf{3. Generation with Multi-Granularity Outputs:}
Using the constructed input (original question + ground truth + 1-shot example), all three teacher models are prompted to generate answers at multiple granularity levels ($G = \{g_1, g_2, \dots, g_6\}$). These levels range from concise summaries to highly detailed, step-by-step reasoning. By solving each question across six levels of granularity, this step systematically evaluates the models’ ability to adapt their reasoning to different levels of abstraction.

\textbf{4. Ranking and Alignment:}
The generated outputs are sorted to align with the original dataset’s order, ensuring consistency and enabling a systematic evaluation of the results. Sorting the generated outputs ensures that the evaluation is systematic and comparable against the original dataset.
\\

\noindent \textbf{Why do we use 1-shot example in the prompt:}

We have decided to incorporate a 1-shot example into the prompt instead of using a 0-shot prompt, based on our trial-and-error findings.

Our initial attempt used a forward-generation approach, where we prompted the model to produce the most succinct response and then enrich it level by level. However, we encountered significant challenges with this approach. The model struggled to demonstrate consistent incremental increases in granularity, as the initial requirement for conciseness often constrained its reasoning and led to inaccuracies or incomplete answers. The model’s inability to build upon a succinct base made this method unsuitable for achieving the desired level of granularity.

To address this, we reversed the approach by asking the model to provide the most elaborate response, intending to progressively reduce the level of detail in subsequent steps. While this method initially produced more detailed outputs, the responses often lacked sufficient depth and structure to support multiple rounds of granularity reduction. As a result, achieving consistent decreases in detail also proved to be a challenge.

These findings highlighted the need for a more structured and balanced approach. We identified that including a 1-shot example in the prompt could effectively guide the model to produce outputs with consistent and balanced granularity across levels. A well-designed 1-shot example helps the model demonstrate high-quality reasoning even in concise answers, ensuring alignment with task requirements regardless of the level of detail. It also provides a clear reference for maintaining consistency when transitioning between levels of granularity.

In summary, 1-shot prompts strike an effective balance between flexibility and structure, enabling the model to generalize across tasks while maintaining coherence and consistency. This approach significantly enhances the model’s ability to generate high-quality training samples with varying levels of reasoning granularity.

As a result, we have decided to generate a 1-shot example to include in the prompt. The first prompt will be used to create the 1-shot example, and the second prompt will leverage it for data generation.

\subsection{Prompts}
\label{sec:appendix_prompt}

\begin{figure*}[h!]
\begin{tcolorbox}[
  colback=gray!10, 
  colframe=black, 
  title=\textbf{CoT Prompt Template}]

You are a math teacher. Please think step by step for the following question.

\vspace{0.5em} 

Output the result strictly in the following format. DO NOT generate any other explanations. 

\vspace{0.5em} 

The generated answer must be consistent with the given answer.

\vspace{0.5em} 

\texttt{Question: "<your question>" Answer: "<original answer>"}

\vspace{0.5em} 

The output format should be as follows:

\texttt{ "instruction": "<your question>", "output": "<Solution Path> The answer is <answer>"}  

\vspace{0.5em} 

Here is the example:

\texttt{<example>}
\end{tcolorbox}
\caption{Prompt for generating CoT dataset.}
\end{figure*}

\begin{figure*}[h!]
\begin{tcolorbox}[
  colback=gray!10, 
  colframe=black, 
  title=\textbf{Synthetic 1-shot CoT Example Based on Granularity Prompt Template}]

You are a math teacher. Please think step by step for the following questions in six different Granularity levels.

\vspace{1em} 

Ensure that the explanations become progressively more detailed as the Granularity increases. The difference in the number of words between each Granularity should be as large as possible.

\vspace{1em} 

The generated answer must be consistent with the given answer. DO NOT generate any other explanations. Output the result strictly in the following format:

\vspace{1em} 

\texttt{"instruction": "<your question>", "output": "<Solution Path>\textbackslash n The answer is <answer>" }

\vspace{1em} 

Granularity definitions:

\vspace{0.3em} 

- \textbf{Level 1}: Provide the most essential steps to reach the answer, minimizing explanations and focusing on the direct path to the solution.

- \textbf{Level 2}: Provide the essential steps required to reach the answer, including some intermediate calculations. It should be more detailed than level 1 but shorter than level 3.

- \textbf{Level 3}: Provide a detailed breakdown that includes all necessary calculations and explanations but shorter and less detailed than level 4. Ensure it is more detailed than level 2.

- \textbf{Level 4}: Provide a very detailed breakdown that includes all necessary calculations and explanations but avoids extra clarifications that would belong to level 5. It should be more detailed than level 3.

- \textbf{Level 5}: Provide an extremely detailed breakdown, covering almost all aspects and sub-steps, but shorter and less detailed than level 6. It should include additional clarifications and checks beyond level 4.

- \textbf{Level 6}: Provide the most detailed and comprehensive breakdown, covering all possible aspects and sub-steps. Ensure this explanation is significantly longer and more detailed than level 5.

\vspace{1em} 

The output format should be as follows:

\vspace{0.3em} 

\texttt{\{ "Granularity": "level 1", "instruction": "<your question>", "output": "Explanation: <Explanation>\textbackslash n The answer is <answer>" \} \\
\{ "Granularity": "level 2", "instruction": "<your question>", "output": "Explanation: <Explanation>\textbackslash n The answer is <answer>" \} \\
\{ "Granularity": "level 3", "instruction": "<your question>", "output": "Explanation: <Explanation>\textbackslash n The answer is <answer>" \} \\
\{ "Granularity": "level 4", "instruction": "<your question>", "output": "Explanation: <Explanation>\textbackslash n The answer is <answer>" \} \\
\{ "Granularity": "level 5", "instruction": "<your question>", "output": "Explanation: <Explanation>\textbackslash n The answer is <answer>" \} \\
\{ "Granularity": "level 6", "instruction": "<your question>", "output": "Explanation: <Explanation>\textbackslash n The answer is <answer>" \}}
\end{tcolorbox}
\caption{Synthetic 1-shot CoT prompt example based on granularity levels, demonstrating progressively detailed reasoning steps from Level 1 to Level 6 explanations.}
\end{figure*}

\clearpage
\subsection{Granularity case study}
\label{appendix:Granularity_case_study}
Question: At the beginning of every period of British Literature, Mrs. Crabapple picks a random student to receive a crabapple as a gift. Given that there are 11 students in her class and her class meets four times a week, how many different sequences of crabapple recipients are possible in a week?
\subsubsection{GPT-4o}
\definecolor{LightGray}{gray}{0.95}
\definecolor{LightCyan}{rgb}{0.88,1,1}

\begingroup
\begin{table*}[h!]
\centering
\setlength{\tabcolsep}{8pt} 
\small
\fontsize{8}{10}\selectfont 
\resizebox{\textwidth}{!}{ 

\begin{tabular}{p{0.15\linewidth} p{0.75\linewidth}} 
\toprule
\rowcolor{LightGray}
\textbf{Granularity} & \textbf{Generated Output} \\ 
\midrule

\rowcolor{white}
Level 1 & 1. Number of students = 11

\vspace{0.5em} 

2. Class meets per week = 4 times

\vspace{0.5em} 

3. Each day, 1 student can receive a crabapple

\vspace{0.5em} 

4. Choices per day = 11

\vspace{0.5em} 

5. Total choices = $11^4 = 14641$

\vspace{0.5em} 

The answer is 14,641 \\

\rowcolor{gray!30}
Level 2 & 1. Each period is an independent random event where 1 of 11 students is chosen.

\vspace{0.5em} 

2. There are 4 periods per week.

\vspace{0.5em} 

3. Each of the 4 periods has 11 possible choices. 

\vspace{0.5em} 

4. The total number of sequences is $11^4$.

\vspace{0.5em} 

5. Calculate $11^4$: $11 \times 11 = 121$, $121 \times 11 = 1331$, $1331 \times 11 = 14641$

\vspace{0.5em} 

The answer is 14,641 \\

\rowcolor{white}
Level 3 & 1. Each time Mrs. Crabapple picks a student, she has 11 choices.

\vspace{0.5em} 

2. Since the class meets four times a week, there are four independent selections, each with 11 choices.

\vspace{0.5em} 

3. The number of different sequences of crabapple recipients is calculated by raising the number of choices to the power corresponding to the number of selections: $11^4$.

\vspace{0.5em} 

4. Calculate this value: $11 \times 11 \times 11 \times 11 = 14,641$.

\vspace{0.5em} 

The answer is 14,641 \\

\rowcolor{gray!30}
Level 4 & 1. Each class session, Mrs. Crabapple picks one of the 11 students.

\vspace{0.5em} 

2. Multiply the number of choices per session: 11 choices per session.

\vspace{0.5em} 

3. Since the class meets four times a week, we need to raise the number of choices to the power of the number of sessions: $11^4$.

\vspace{0.5em} 

4. Calculate $11^4$: $11 \times 11 \times 11 \times 11$.

\vspace{0.5em} 

5. Perform the multiplication step-by-step: $11 \times 11 = 121$, $121 \times 11 = 1331$, $1331 \times 11 = 14,641$.

\vspace{0.5em} 

6. Review all steps for consistency: 11 choices per session, 4 sessions total, $11^4$ computation confirms 14,641 possibilities.

\vspace{0.5em} 

The answer is 14,641 \\

\rowcolor{white}
Level 5 & 1. Identify key variables: 11 students, 4 class meetings.

\vspace{0.5em} 

2. Recognize the problem's nature: finding the number of different sequences (order matters) of recipients over 4 periods.

\vspace{0.5em} 

3. Analyze each period's choice: Each of the 4 periods can be filled by any one of the 11 students.

\vspace{0.5em} 

4. Calculate total sequences using the multiplication principle: Each of 4 periods has 11 possible choices.

\vspace{0.5em} 

5. Compute total different sequences: $11 \times 11 \times 11 \times 11 = 11^4$.

\vspace{0.5em} 

6. Evaluate numerical result: $11^4 = 11 \times 11 \times 11 \times 11 = 14,641$.

\vspace{0.5em} 

The answer is 14,641 \\

\rowcolor{gray!30}
Level 6 & 1. Determine possible choices per class: Mrs. Crabapple can choose one out of 11 students each time.

\vspace{0.5em} 

2. Confirm weekly meetings: The class has four periods a week.

\vspace{0.5em} 

3. Establish sequence combination formula: For each period, there are 11 choices, resulting in 11 choices for the first period, 11 for the second, 11 for the third, and 11 for the fourth.

\vspace{0.5em} 

4. Calculate total sequences: Multiply available choices for each period to find total combinations: $11 \times 11 \times 11 \times 11 = 11^4$.

\vspace{0.5em} 

5. Compute final result: Raising 11 to the power of 4, we get $11^4 = 14,641$ sequences.

\vspace{0.5em} 

The answer is 14,641 \\

\bottomrule
\end{tabular}
}
\caption{Comparison of Different CoT granularity for the same problem generated by GPT-4o.} 

\end{table*}

\endgroup

\newpage
\subsubsection{LLaMA 3 70B}
\definecolor{LightGray}{gray}{0.95}
\definecolor{LightCyan}{rgb}{0.88,1,1}

\begingroup
\renewcommand{\arraystretch}{1.4} 

\begin{table*}[h!]
\centering
\setlength{\tabcolsep}{8pt}
\fontsize{8}{10}\selectfont 

\begin{tabular}{p{0.12\linewidth} p{0.78\linewidth} r} 
\toprule
\rowcolor{LightGray}
\textbf{Granularity} & \textbf{Generated Output} \\ 
\midrule

\rowcolor{white}
Level 1 & The number of possible recipients for each class is 11.

\vspace{0.5em} 

Since class meets 4 times a week, the total number of sequences is 11*11*11*11 = $11^4$ = 14641.

\vspace{0.5em}  

The answer is 14,641 \\

\rowcolor{gray!30}
Level 2 & First, we need to find the number of possible recipients for each class period, which is 11.

\vspace{0.5em} 

Then, we need to find the number of possible sequences for the week, which is the number of possible recipients for each class period raised to the power of the number of class periods in a week: $11^4$ = 14641.

\vspace{0.5em} 

The answer is 14,641 \\

\rowcolor{white}
Level 3 & Step 1: Determine the number of students in the class. There are 11 students in Mrs. Crabapple\'s British Literature class. 

\vspace{0.5em} 

Step 2: Determine the number of times the class meets in a week. The class meets 4 times a week. 

\vspace{0.5em} 

Step 3: Calculate the number of different sequences of crabapple recipients possible in a week. Since there are 11 students and the class meets 4 times a week, we can calculate the number of different sequences by raising 11 to the power of 4, which represents the number of choices for each of the 4 days. $11^4$ = 14641 

\vspace{0.5em} 

The answer is 14,641 \\

\rowcolor{gray!30}
Level 4 & Step 1: Determine the number of students in the class. There are 11 students in Mrs. Crabapple\'s British Literature class. 

\vspace{0.5em} 

Step 2: Determine the number of times the class meets in a week. The class meets 4 times a week. 

\vspace{0.5em} 

Step 3: Calculate the number of different sequences of crabapple recipients possible in a week. Since Mrs. Crabapple picks a random student each time, the number of sequences is equal to the number of permutations of 11 students taken 4 at a time, allowing for repetition (since the same student can be picked more than once). This is calculated as $11^4$.

\vspace{0.5em} 

Step 4: Calculate $11^4$. $11^4$ = 11 * 11 * 11 * 11 = 14641. 

\vspace{0.5em} 

The answer is 14,641 \\

\rowcolor{white}
Level 5 & To solve this problem, we need to use the concept of permutations. 

\vspace{0.5em} 

Since Mrs. Crabapple picks a random student each time, the order of selection matters. We have 11 students to choose from for each of the 4 periods in a week. 

\vspace{0.5em} 

For the first period, there are 11 possible choices. For the second period, there are again 11 possible choices, and so on. Since the choices are independent of each other, we can multiply the number of choices for each period to get the total number of different sequences. This gives us (11 times 11 times 11 times 11 = $11^4$ = 14641). 

\vspace{0.5em} 

Therefore, there are 14641 different sequences of crabapple recipients possible in a week.

\vspace{0.5em} 

The answer is 14,641 \\

\rowcolor{gray!30}
Level 6 & To solve this problem, we need to use the concept of permutations. 

\vspace{0.5em} 

Since Mrs. Crabapple picks a random student to receive a crabapple as a gift in each class, and there are 11 students in her class, there are 11 possible choices for the first class.For the second class, there are also 11 possible choices, and so on. 

\vspace{0.5em} 

Since her class meets four times a week, we need to find the total number of different sequences of crabapple recipients for four classes. This can be calculated by multiplying the number of choices for each class together. 

\vspace{0.5em} 

So, the total number of different sequences is (11 times 11 times 11 times 11 = $11^4$). To calculate ($11^4$), we can multiply 11 by itself four times.

\vspace{0.5em} 

This gives us 11 times 11 = 121, 121 times 11 = 1331, and 1331 times 11 = 14641. Therefore, the total number of different sequences of crabapple recipients possible in a week is 14641.

\vspace{0.5em} 

The answer is 14,641 \\

\bottomrule
\end{tabular}
\caption{Comparison of Different CoT granularity for the same problem generated by LLaMA 3 70B.} 

\end{table*}
\endgroup

\newpage
\subsubsection{Gemini-1.5-Flash}

\definecolor{LightGray}{gray}{0.95}
\definecolor{LightCyan}{rgb}{0.88,1,1}

\begingroup
\renewcommand{\arraystretch}{1.4} 

\begin{table*}[h!]
\centering
\setlength{\tabcolsep}{8pt} 
\fontsize{8}{10}\selectfont 

\begin{tabular}{p{0.12\linewidth} p{0.78\linewidth} r} 
\toprule
\rowcolor{LightGray}
\textbf{Granularity} & \textbf{Generated Output} \\ 
\midrule

\rowcolor{white}
Level 1 & There are 11 choices for the recipient of the crabapple in each of the four class periods.   

\vspace{0.5em} 

The number of possible sequences is therefore 11 * 11 * 11 * 11 = $11^4$. $11^4$ = 14641.

\vspace{0.5em}  

The answer is 14,641 \\

\rowcolor{gray!30}
Level 2 & There are 11 students in the class.

\vspace{0.5em} 

Mrs. Crabapple picks a student each of the four class periods. The number of choices for the first period is 11. The number of choices for the second period is 11. The number of choices for the third period is 11. The number of choices for the fourth period is 11. 

\vspace{0.5em} 

To find the total number of different sequences, we multiply the number of choices for each period together: 11 * 11 * 11 * 11 = $11^4$ = 14641. 

\vspace{0.5em} 

The answer is 14,641 \\

\rowcolor{white}
Level 3 & There are 11 students in the class.

\vspace{0.5em} 

Mrs. Crabapple picks a student each of the four class periods. The number of choices for the first period is 11. The number of choices for the second period is also 11 (since the same student can receive a crabapple multiple times). Similarly, there are 11 choices for the third period and 11 choices for the fourth period.  

\vspace{0.5em} 

To find the total number of different sequences, we multiply the number of choices for each period together. This is 11 times 11 times 11 times 11 = $11^4$.
Calculating $11^4$ gives us 14641.

\vspace{0.5em} 

The answer is 14,641 \\

\rowcolor{gray!30}
Level 4 & There are 11 students in the class, and the class meets 4 times a week.

\vspace{0.5em} 

Each time Mrs. Crabapple picks a student, there are 11 choices.  

\vspace{0.5em} 

Since there are four periods, we multiply the number of choices for each period together to find the total number of different sequences. This is 11 * 11 * 11 * 11 = $11^4$. Calculating $11^4$: 11 * 11 = 121; 121 * 11 = 1331; 1331 * 11 = 14641. 

\vspace{0.5em} 

The answer is 14,641 \\

\rowcolor{white}
Level 5 & There are 11 students in the class, and the class meets four times a week.

\vspace{0.5em} 

Each time the class meets, Mrs. Crabapple picks one student to receive a crabapple.  The number of choices for the first period is 11.  The number of choices for the second period is also 11 (since the same student can receive a crabapple again).  Similarly, there are 11 choices for the third period and 11 choices for the fourth period. 

\vspace{0.5em} 

To find the total number of different sequences of crabapple recipients, we multiply the number of choices for each period together. This is because each choice is independent of the others. Therefore, the total number of sequences is 11 * 11 * 11 * 11 = $11^4$. 

\vspace{0.5em} 

Calculating $11^4$: 11 * 11 = 121; 121 * 11 = 1331; 1331 * 11 = 14641. 

\vspace{0.5em} 

The answer is 14,641 \\

\rowcolor{gray!30}
Level 6 & There are 11 students in the class, and the class meets four times a week.  

\vspace{0.5em} 

Each time the class meets, Mrs. Crabapple picks one student to receive a crabapple. The number of ways to choose a student for the first period is 11.  The number of ways to choose a student for the second period is also 11.  Similarly, there are 11 ways to choose a student for the third period and 11 ways to choose a student for the fourth period. 

\vspace{0.5em} 

Since these choices are independent, we multiply the number of choices for each period together to find the total number of different sequences of crabapple recipients in a week. This is 11 * 11 * 11 * 11 = $11^4$. 

\vspace{0.5em} 

Calculating $11^4$, we get 14641. Therefore, there are 14,641 different sequences of crabapple recipients possible in a week. 

\vspace{0.5em} 

The answer is 14,641 \\

\bottomrule
\end{tabular}
\caption{Comparison of Different CoT granularity for the same problem generated by Gemini-1.5-Flash.} 

\end{table*}
\endgroup

\newpage
\section{Different CoT Format Dataset Collection}
\label{appendix:Format}
\subsection{Workflow} 
\textbf{The dataset with format processing steps are detailed below:}  

\textbf{1. 0-Shot Example Generation:}  
The same question is first provided to ChatGPT using a 0-shot prompt in three formats. ChatGPT generates a single 1-shot example for each format (including both the question and a corresponding output in the specific format). This step ensures that ChatGPT establishes a baseline example for each format. These generated examples serve as templates for subsequent responses, ensuring consistency in style and logic.  

\textbf{2. Input Construction:}  
Each question is then re-input into ChatGPT, along with:  
- Its corresponding original ChatGPT-generated outputs.  
- The 1-shot examples generated in Step 1 for all three reasoning formats.  

Including the 1-shot examples in the input serves as explicit format demonstrations, guiding ChatGPT to generate outputs that align with the desired styles. This process improves the coherence and quality of the resulting outputs.  

\textbf{3. Multi-Format Output Generation:}  
Using the constructed input (original question + original outputs + 1-shot examples), ChatGPT generates reformatted outputs for each question across three reasoning formats while preserving the original logic: Least-to-most, RaR and SymbolicCoT.  

\textbf{4. Ranking and Alignment:}  
The reformatted outputs are then sorted to align with the original dataset’s order. This step ensures consistency and enables systematic evaluation. Sorting the outputs guarantees that the evaluation is both structured and comparable across different reasoning formats and the original dataset.

\subsection{Prompt}

\begin{figure*}[h!]
\small
\begin{tcolorbox}[
  colback=gray!10, 
  colframe=black, 
  title=\textbf{Symbolic CoT Prompt Template}]

Please rewrite the output by following the Symbolic CoT (SymbCoT) reasoning to solve the given question step-by-step. You can ONLY change the format but not the original steps. In your rewrite, translate the question’s context into symbolic logic format, identifying key variables and relationships. Ensure to use logical symbols such as $\exists$ (exists), $\forall$ (for all), $\land$ (and), $\lor$ (or) and $\implies$ (implies), etc., to represent relationships between variables.

\noindent You should use symbolic thinking steps in the output.  
The generated output must follow this specific structure and include logical symbols. Output the result strictly in the following format. DO NOT generate any other explanations.  

The generated answer must be consistent with the given answer. After modification, you must add \texttt{The answer is <answer>} at the end.  

\noindent Here is the original output:  
\[
\{ \texttt{instruction: <question>}, \texttt{output: <solution\_path>} \}
\]

\noindent The output format should be as follows:  
\[
\{ \texttt{instruction: <question>}, \texttt{output: <SymbCoT Solution Path> The answer is <answer>} \}  
\]

\noindent Here is the example:  
\texttt{<example>}

\end{tcolorbox}
\caption{Prompt for generating Symbolic Chain-of-Thought (SymbCoT) reasoning, requiring the transformation of problem contexts into symbolic logic representations using logical operators ($\exists$, $\forall$, $\land$, $\lor$ and $\implies$). }
\end{figure*}

\begin{figure*}[h!]
\small
\begin{tcolorbox}[
  colback=gray!10, 
  colframe=black, 
  title=\textbf{Rephrase and Respond (RaR) CoT Prompt Template}]

Please rewrite the output and answer them individually. You can ONLY change the format but not the original steps. 
The generated answer must be consistent with the given answer. Output the result strictly in the following format. DO NOT generate any other explanations.  

After modification, you must add \texttt{The answer is <answer>} at the end.  

\noindent Here is the original output:  
\[
\{ \texttt{instruction: <question>}, \texttt{output: <solution\_path>} \}
\]

\noindent Rephrase and expand the given question, and then respond carefully.  

\noindent The output format should be as follows:  
\[
\begin{aligned}
    &\{ \texttt{instruction: <question>}, \\
    &\quad \texttt{output: <Rephrase and expand the given question>} \\
    &\quad \texttt{<solution path> The answer is <answer>} \}
\end{aligned}
\]

\noindent Here is the example:  
\texttt{<example>}

\end{tcolorbox}
\caption{Prompt for Rephrase and Respond (RaR), which restructures the question while maintaining the original reasoning steps. }
\end{figure*}

\begin{figure*}[h!]
\small
\begin{tcolorbox}[
  colback=gray!10, 
  colframe=black, 
  title=\textbf{Least-to-Most CoT Prompt Template}]

Please rewrite the output by generating sub-questions and answer them individually. You can ONLY change the format but not the original steps.  

The generated answer must be consistent with the given answer. Output the result strictly in the following format. DO NOT generate any other explanations.  

After modification, you must add \texttt{The answer is <answer>} at the end.  

\noindent Here is the original output:  
\[
\{ \texttt{instruction: <question>}, \texttt{output: <solution\_path>} \}  
\]

\noindent The output format should be as follows:  
\[
\begin{aligned}
    &\{ \texttt{instruction: <question>}, \\
    &\quad \texttt{output: <Subquestion\_1><Explanation\_1>} \\
    &\quad \texttt{<Subquestion\_2><Explanation\_2>... The answer is <answer>} \}
\end{aligned}
\]

\noindent Here is the example:  
\texttt{<example>}

\end{tcolorbox}
\caption{Prompt for Least-to-Most CoT, which restructures the solution by generating and answering sub-questions step-by-step while preserving the original reasoning process.}
\end{figure*}

\clearpage
\subsection{Format Case Study}

\begin{figure*}[h!]
\begin{tcolorbox}[colback=gray!10, colframe=black, title=Different CoT Format Examples, center title]
\small

\begin{minipage}{\textwidth}
\begin{minipage}[t]{0.48\textwidth} 
\textbf{Original CoT:}

Bert currently has 20 kangaroos. Kameron has 100 kangaroos. The difference in the number of kangaroos is 100 - 20 = 80 kangaroos. Bert buys kangaroos at a rate of 2 kangaroos per day. Therefore, the number of days required for Bert to have the same number of kangaroos as Kameron currently has is 80 / 2 = 40 days. The answer is 40.
\end{minipage}
\hfill
\vrule
\hfill
\begin{minipage}[t]{0.48\textwidth} 
\textbf{Symbolic CoT:}

1. Let \( K_0 = 100 \) \( \exists K_0 \). (Kameron has 100 kangaroos).  

2. Let \( B_0 = 20 \) \( \exists B_0 \). (Bert has 20 kangaroos).  

3. The difference in kangaroos, \( D = K_0 - B_0 \) \( \exists D \):  
   \[
   D = 100 - 20 = 80
   \]

4. Rate of buying kangaroos per day, \( r = 2 \) \( \forall r \):  
   \[
   r = 2
   \]

5. Number of days required, \( N = \frac{D}{r} \) \( \exists N \):  
   \[
   N = \frac{80}{2} = 40
   \]

The answer is 40.
\end{minipage}
\end{minipage}

\vspace{1em}
\hrule
\vspace{1em}

\begin{minipage}{\textwidth}
\begin{minipage}[t]{0.48\textwidth} 
\textbf{Rephrase and Respond:}

\textcolor{darkgreen}{On Kameron's large farm, there are currently 100 kangaroos, whereas Bert's farm has only 20 kangaroos. Bert aims to increase his kangaroo population, purchasing 2 new kangaroos each day, until he reaches the current number of kangaroos on Kameron's farm.}

Calculate the number of days Bert will need to achieve this goal. Bert currently has 20 kangaroos. Kameron has 100 kangaroos. The difference in the number of kangaroos is 100 - 20 = 80 kangaroos. Bert buys kangaroos at a rate of 2 kangaroos per day. Therefore, the number of days required for Bert to have the same number of kangaroos as Kameron currently has is 80 / 2 = 40 days. The answer is 40.
\end{minipage}
\hfill
\vrule
\hfill
\begin{minipage}[t]{0.48\textwidth} 
\textbf{\textsc{Least-to-most}:}

\textcolor{darkgreen}{How many kangaroos does Bert currently have?}  
Bert currently has 20 kangaroos.  
\textcolor{darkgreen}{How many kangaroos does Kameron currently have?}  
Kameron has 100 kangaroos.  
\textcolor{darkgreen}{What is the difference in the number of kangaroos between Bert and Kameron?}  
The difference in the number of kangaroos is 100 - 20 = 80 kangaroos.  
\textcolor{darkgreen}{What is the rate at which Bert buys kangaroos per day?}  
Bert buys kangaroos at a rate of 2 kangaroos per day.  
\textcolor{darkgreen}{How many days will it take for Bert to have the same number of kangaroos as Kameron?}  
The number of days required for Bert to have the same number of kangaroos as Kameron currently has is 80 / 2 = 40 days.  

The answer is 40.
\end{minipage}
\end{minipage}

\end{tcolorbox}
\caption{Comparison of Different CoT Formats for the same problem: \textbf{Kameron has 100 kangaroos on his large farm; Bert has 20 kangaroos on his farm. In how many more days will Bert have the same number of kangaroos as Kameron does now if he buys kangaroos at the same rate of 2 new kangaroos per day?}}
\end{figure*}

\clearpage
\newpage
\section{Whole Results of Granularity Experiments} 
\label{appendix:Granularityresults}

\begin{table*}[h]
\centering
\small
\renewcommand{\arraystretch}{0.7}
\resizebox{0.90\textwidth}{!}{
\begin{tabular}{>{\centering\arraybackslash}m{2.5cm} |c|llllll}
\toprule
\multirow{2}{*}{\textbf{Dataset}} & \multirow{2}{*}{\textbf{Only Answer}} & \multicolumn{6}{c}{\textbf{Gemma 2B Performance}} \\ 
\cmidrule(lr){3-8}
& & Level 1 & Level 2 & Level 3 & Level 4 & Level 5 & Level 6 \\
\midrule
SVAMP & 47.70 & 59\textsubscript{±4.58} & 64.33\textsubscript{±0.00} & 65.22\textsubscript{±0.69} & 65.89\textsubscript{±0.38} & \textbf{67.11\textsubscript{±1.35}}$^{\color{red}{\uparrow\scriptsize13.74\%}}$ & 66.89\textsubscript{±1.02} \\
GSM8K & 8.20 & 49.66\textsubscript{±0.27} & 52.36\textsubscript{±0.98} & 53.37\textsubscript{±0.33} & 52.69\textsubscript{±0.13} & 53.42\textsubscript{±0.83}& \textbf{53.45\textsubscript{±1.48}}$^{\color{red}{\uparrow\scriptsize7.63\%}}$  \\
AQuA-RAT & 20.47 & 40.68\textsubscript{±1.27} & 42.91\textsubscript{±1.42} & 43.7\textsubscript{±2.58} & 39.9\textsubscript{±1.49} & \textbf{44.88\textsubscript{±0.79}}$^{\color{red}{\uparrow\scriptsize12.48\%}}$ & 44.49\textsubscript{±2.36} \\
MATH & 9.00 & 23.4\textsubscript{±1.06} & 21.53\textsubscript{±2.16} & \textbf{24.4\textsubscript{±0.20}}$^{\color{red}{\uparrow\scriptsize16.19\%}}$ & 21.93\textsubscript{±0.42} & 23.0\textsubscript{±1.22} & 21.0\textsubscript{±0.69} \\
CSQA & \textbf{69.86} & 67.38\textsubscript{±0.82} & 67.98\textsubscript{±0.37} & 
68.74\textsubscript{±1.30} & 
66.75\textsubscript{±0.53} & 67.54\textsubscript{±0.47} & 66.01\textsubscript{±1.50} \\
OBQA & 69.60 & 71.53\textsubscript{±1.94} & 69.93\textsubscript{±0.90} & 69.93\textsubscript{±1.36} & 68.33\textsubscript{±1.27} & \textbf{72.00\textsubscript{±1.64}}$^{\color{red}{\uparrow\scriptsize5.37\%}}$ & 70.13\textsubscript{±1.62} \\
STQA & 60.69 & \textbf{67.59\textsubscript{±1.04}}$^{\color{red}{\uparrow\scriptsize7.11\%}}$ & 63.1\textsubscript{±1.79} & 64.6\textsubscript{±1.56} & 63.45\textsubscript{±1.24} & 65.75\textsubscript{±1.77} & 64.14\textsubscript{±1.58} \\
\midrule
\multirow{2}{*}{\textbf{Dataset}} & \multirow{2}{*}{\textbf{Only Answer}} & \multicolumn{6}{c}{\textbf{LLaMA 3.2 1B Performance}} \\ 
\cmidrule(lr){3-8}
& & Level 1 & Level 2 & Level 3 & Level 4 & Level 5 & Level 6 \\
\midrule
SVAMP & 37.70 & 52.67\textsubscript{±2.52} & 52.11\textsubscript{±2.27} & 52.78\textsubscript{±0.84} & \textbf{53.44\textsubscript{±1.17}}$^{\color{red}{\uparrow\scriptsize2.55\%}}$ & 52.44\textsubscript{±1.71} & 52.78\textsubscript{±3.5} \\
GSM8K & 6.70 & 36.8\textsubscript{±0.77} & 39.73\textsubscript{±0.67} & \textbf{40.08\textsubscript{±0.98}$^{\color{red}{\uparrow\scriptsize8.91\%}}$} & 39.32\textsubscript{±1.33} & 39.58\textsubscript{±1.04} & 38.54\textsubscript{±0.96} \\
AQuA-RAT & 24.00 & \textbf{34.12\textsubscript{±1.82}}$^{\color{red}{\uparrow\scriptsize12.57\%}}$ & 30.31\textsubscript{±1.42} & 30.58\textsubscript{±1.2} & 31.23\textsubscript{±2.02} & 33.2\textsubscript{±2.17} & 30.45\textsubscript{±0.91} \\
MATH & 7.00 & \textbf{8.87\textsubscript{±0.92}}$^{\color{red}{\uparrow\scriptsize11.85\%}}$ & 8.07\textsubscript{±1.10} & 8.4\textsubscript{±0.35} & 8.27\textsubscript{±0.12} & 7.93\textsubscript{±1.14} & 8.33\textsubscript{±0.12} \\ 
CSQA & 19.57 & \textbf{64.48\textsubscript{±1.20}}$^{\color{red}{\uparrow\scriptsize5.39\%}}$ & 63.25\textsubscript{±0.34} & 62.9\textsubscript{±1.21} & 61.94\textsubscript{±1.58} & 62.68\textsubscript{±1.18} & 61.18\textsubscript{±0.41} \\ 
OBQA & 51.60 & \textbf{64.4\textsubscript{±1.25}}$^{\color{red}{\uparrow\scriptsize2.01\%}}$ & 63.73\textsubscript{±1.36} & 63.6\textsubscript{±2.12} & 63.6\textsubscript{±1.25} & 63.27\textsubscript{±1.36} & 63.13\textsubscript{±0.70} \\ 
STQA & 53.10 & 63.33\textsubscript{±1.59} & 60.11\textsubscript{±1.30} & 63.56\textsubscript{±1.59} & \textbf{64.14\textsubscript{±1.50}}$^{\color{red}{\uparrow\scriptsize6.70\%}}$ & 61.84\textsubscript{±1.44} & 63.33\textsubscript{±2.30} \\ 
\midrule
\multirow{2}{*}{\textbf{Dataset}} & \multirow{2}{*}{\textbf{Only Answer}} & \multicolumn{6}{c}{\textbf{LLaMA 3.2 3B Performance}} \\ 
\cmidrule(lr){3-8}
& & Level 1 & Level 2 & Level 3 & Level 4 & Level 5 & Level 6 \\
\midrule
SVAMP & 53.70 & 69\textsubscript{±3.61} & 65.89\textsubscript{±3.53} & 68.33\textsubscript{±1.86} & 68.11\textsubscript{±3.27} & 69.78\textsubscript{±1.07} & \textbf{74.33\textsubscript{±1.45}}$^{\color{red}{\uparrow\scriptsize12.81\%}}$ \\ 
GSM8K & 9.30 & 59.59\textsubscript{±1.59} & 62.29\textsubscript{±1.41} & 62.57\textsubscript{±0.80} & \textbf{63.48\textsubscript{±0.16}}$^{\color{red}{\uparrow\scriptsize6.53\%}}$ & 62.29\textsubscript{±1.22} & 60.98\textsubscript{±0.31} \\ 
AQuA-RAT & 19.60 & 44.36\textsubscript{±2.31} & 44.88\textsubscript{±2.19} & 45.01\textsubscript{±2.37} & 46.19\textsubscript{±3.94} & \textbf{47.24\textsubscript{±4.77}}$^{\color{red}{\uparrow\scriptsize6.49\%}}$ & 46.33\textsubscript{±3.01} \\ 
MATH & 9.40 & 19.07\textsubscript{±0.90} & 19.6\textsubscript{±1.06} & 19.73\textsubscript{±1.72} & \textbf{20.27\textsubscript{±1.42}}$^{\color{red}{\uparrow\scriptsize11.37\%}}$ & 19.93\textsubscript{±2.20} & 18.2\textsubscript{±1.64} \\ 
CSQA & 62.00 & 72.62\textsubscript{±0.82} & 70.71\textsubscript{±0.70} & \textbf{74.12\textsubscript{±0.50}}$^{\color{red}{\uparrow\scriptsize4.82\%}}$ & 71.75\textsubscript{±0.62} & 71.17\textsubscript{±1.03} & 71.44\textsubscript{±0.90} \\ 
OBQA & 74.40 & 79.33\textsubscript{±0.42} & 79.73\textsubscript{±0.70} & 78.8\textsubscript{±0.80} & 77.8\textsubscript{±0.92} & 79.27\textsubscript{±2.04} & \textbf{80.2\textsubscript{±1.78}}$^{\color{red}{\uparrow\scriptsize3.08\%}}$ \\ 
STQA & 55.52 & 66.44\textsubscript{±1.55} & 62.76\textsubscript{±2.82} & 67.47\textsubscript{±1.44} & 66.78\textsubscript{±1.39} & 63.91\textsubscript{±2.79} & \textbf{68.62\textsubscript{±1.20}}$^{\color{red}{\uparrow\scriptsize9.34\%}}$ \\ 
\midrule
\multirow{2}{*}{\textbf{Dataset}} & \multirow{2}{*}{\textbf{Only Answer}} & \multicolumn{6}{c}{\textbf{BLOOM 560M Performance}} \\ 
\cmidrule(lr){3-8}
& & Level 1 & Level 2 & Level 3 & Level 4 & Level 5 & Level 6 \\
\midrule
SVAMP & 0.00 & 5.11\textsubscript{±0.19} & 4.56\textsubscript{±0.77} & \textbf{6.67\textsubscript{±1.33}}$^{\color{red}{\uparrow\scriptsize46.27\%}}$ & 6.56\textsubscript{±1.90} & 5.56\textsubscript{±2.41} & 5.11\textsubscript{±0.69} \\ 
GSM8K & 3.90 & 7.25\textsubscript{±0.77} & 8.11\textsubscript{±0.08} & \textbf{8.47\textsubscript{±0.12}}$^{\color{red}{\uparrow\scriptsize16.83\%}}$ & 7.73\textsubscript{±0.82} & 8.19\textsubscript{±0.27} & 8.11\textsubscript{±0.77} \\ 
AQuA-RAT & 20.90 & \textbf{22.05\textsubscript{±3.43}}$^{\color{red}{\uparrow\scriptsize18.29\%}}$ & 20.6\textsubscript{±2.56} & 21.13\textsubscript{±4.55} & 21.52\textsubscript{±1.59} & 18.64\textsubscript{±1.98} & 19.69\textsubscript{±1.04} \\ 
MATH & \textbf{4.00} & 2.6\textsubscript{±0.53} & 2.33\textsubscript{±1.01} & 2.13\textsubscript{±0.61} & 2.13\textsubscript{±1.15} & 1.67\textsubscript{±0.42} & 1.87\textsubscript{±0.50} \\ 
CSQA & 20.15 & 37.76\textsubscript{±1.29} & \textbf{37.95\textsubscript{±0.29}}$^{\color{red}{\uparrow\scriptsize18.02\%}}$ & 37.84\textsubscript{±0.74} & 33.99\textsubscript{±0.83} & 33.96\textsubscript{±1.57} & 31.89\textsubscript{±1.23} \\ 
OBQA & 34.00 & \textbf{41.07\textsubscript{±0.95}}$^{\color{red}{\uparrow\scriptsize18.04\%}}$ & 38.27\textsubscript{±0.42} & 36.73\textsubscript{±0.76} & 35.73\textsubscript{±1.40} & 34.8\textsubscript{±3.64} & 36.27\textsubscript{±2.04} \\ 
STQA & \textbf{56.90} & 53.1\textsubscript{±1.25} & 52.18\textsubscript{±1.30} & 52.07\textsubscript{±0.35} & 52.99\textsubscript{±0.40} & 53.45\textsubscript{±1.58} & 54.83\textsubscript{±1.73} \\ 
\midrule
\multirow{2}{*}{\textbf{Dataset}} & \multirow{2}{*}{\textbf{Only Answer}} & \multicolumn{6}{c}{\textbf{BLOOM 1.1B Performance}} \\ 
\cmidrule(lr){3-8}
& & Level 1 & Level 2 & Level 3 & Level 4 & Level 5 & Level 6 \\
\midrule
SVAMP & 1.00 & 7.22\textsubscript{±0.69} & 9.11\textsubscript{±0.51} & \textbf{13.00\textsubscript{±1.86}}$^{\color{red}{\uparrow\scriptsize80.06\%}}$ & 11.33\textsubscript{±1.20} & 10.67\textsubscript{±1.00} & 9.89\textsubscript{±1.64} \\ 
GSM8K & 4.20 & 11.32\textsubscript{±0.16} & 14.3\textsubscript{±0.37} & \textbf{14.33\textsubscript{±0.79}}$^{\color{red}{\uparrow\scriptsize26.59\%}}$ & 14.28\textsubscript{±0.66} & 13.09\textsubscript{±0.83} & 11.78\textsubscript{±0.77} \\ 
AQuA-RAT & 22.40 & 21.39\textsubscript{±2.27} & 21.78\textsubscript{±1.20} & 21.92\textsubscript{±1.94} & 21.78\textsubscript{±1.27} & 21.92\textsubscript{±3.66} & \textbf{23.36\textsubscript{±3.16}}$^{\color{red}{\uparrow\scriptsize9.21\%}}$ \\ 
MATH & \textbf{3.20} & 2.33\textsubscript{±0.31} & 2.73\textsubscript{±0.31} & 2.4\textsubscript{±0.40} & 2.07\textsubscript{±0.58} & 2.2\textsubscript{±1.22} & 2.87\textsubscript{±0.76} \\ 
CSQA & 41.69 & \textbf{49.14\textsubscript{±2.21}}$^{\color{red}{\uparrow\scriptsize18.35\%}}$ & 48.21\textsubscript{±2.35} & 48.57\textsubscript{±0.36} & 43.87\textsubscript{±0.82} & 45.37\textsubscript{±0.79} & 41.52\textsubscript{±0.91} \\ 
OBQA & \textbf{48.60} & 48.4\textsubscript{±3.30} & 47.27\textsubscript{±1.81} & 46.07\textsubscript{±2.23} & 46.00\textsubscript{±2.09} & 46.2\textsubscript{±2.62} & 46.4\textsubscript{±2.96} \\ 
STQA & 58.97
& 57.93\textsubscript{±2.39} & \textbf{59.65\textsubscript{±2.48} }$^{\color{red}{\uparrow\scriptsize6.35\%}}$& 58.05\textsubscript{±2.08} & 58.97\textsubscript{±2.41} & 56.09\textsubscript{±2.22} & 57.36\textsubscript{±1.21} \\ 
\midrule
\multirow{2}{*}{\textbf{Dataset}} & \multirow{2}{*}{\textbf{Only Answer}} & \multicolumn{6}{c}{\textbf{BLOOM 1.7B Performance}} \\ 
\cmidrule(lr){3-8}
& & Level 1 & Level 2 & Level 3 & Level 4 & Level 5 & Level 6 \\
\midrule
SVAMP & 0.00 & 10\textsubscript{±1.67} & 9.44\textsubscript{±1.02} & \textbf{17.11\textsubscript{±1.17}}$^{\color{red}{\uparrow\scriptsize71.1\%}}$ & 16.44\textsubscript{±1.39} & 16.56\textsubscript{±0.51} & 11.33\textsubscript{±2.91} \\ 
GSM8K & 5.10 & 13.12\textsubscript{±0.33} & 17.11\textsubscript{±0.62} & 16.68\textsubscript{±0.46} & \textbf{17.89\textsubscript{±1.25}}$^{\color{red}{\uparrow\scriptsize36.36\%}}$ & 16.86\textsubscript{±1.25} & 15.21\textsubscript{±0.12} \\ 
AQuA-RAT & 23.60 & \textbf{25.07\textsubscript{±0.91}}$^{\color{red}{\uparrow\scriptsize14.37\%}}$ & 23.88\textsubscript{±0.45} & 22.05\textsubscript{±5.46} & 22.7\textsubscript{±0.82} & 22.31\textsubscript{±1.38} & 21.92\textsubscript{±2.17} \\ 
MATH & \textbf{3.60}
& 2.87\textsubscript{±0.50} & 1.9\textsubscript{±0.14} & 2.8\textsubscript{±0.53} & 1.87\textsubscript{±0.12} & 2.0\textsubscript{±0.60} & 2.6\textsubscript{±0.53} \\ 
CSQA & 21.38 & \textbf{53.37\textsubscript{±2.21}}$^{\color{red}{\uparrow\scriptsize15.12\%}}$ & 51.24\textsubscript{±0.90} & 51.52\textsubscript{±0.62} & 49.06\textsubscript{±0.59} & 46.79\textsubscript{±0.68} & 46.36\textsubscript{±0.59} \\ 
OBQA & 48.20 & 49.93\textsubscript{±3.00} & \textbf{50.93\textsubscript{±0.99}}$^{\color{red}{\uparrow\scriptsize8.99\%}}$ & 48.0\textsubscript{±1.40} & 47.6\textsubscript{±1.00} & 47.67\textsubscript{±2.50} & 46.73\textsubscript{±2.37} \\ 
STQA & \textbf{58.62}
& 54.94\textsubscript{±1.55} & 56.44\textsubscript{±2.49} & 58.28\textsubscript{±0.91} & 57.24\textsubscript{±2.69} & 56.89\textsubscript{±5.10} & 56.44\textsubscript{±1.90} \\ 
\midrule
\multirow{2}{*}{\textbf{Dataset}} & \multirow{2}{*}{\textbf{Only Answer}} & \multicolumn{6}{c}{\textbf{BLOOM 3B Performance}} \\ 
\cmidrule(lr){3-8}
& & Level 1 & Level 2 & Level 3 & Level 4 & Level 5 & Level 6 \\
\midrule
SVAMP & 5.00 & 15.44\textsubscript{±0.51} & 23.67\textsubscript{±0.00} & 23.11\textsubscript{±1.26} & \textbf{24.00\textsubscript{±0.67}}$^{\color{red}{\uparrow\scriptsize55.44\%}}$ & 22.22\textsubscript{±0.69} & 22.22\textsubscript{±1.02} \\ 
GSM8K & 4.60 & 18.2\textsubscript{±0.57} & 22.34\textsubscript{±1.14} & \textbf{23.81\textsubscript{±0.65}}$^{\color{red}{\uparrow\scriptsize30.82\%}}$ & 22.57\textsubscript{±0.88} & 22.47\textsubscript{±0.86} & 20.85\textsubscript{±0.15} \\ 
AQuA-RAT & \textbf{28.00}
& 24.67\textsubscript{±0.82} & 24.41\textsubscript{±1.72} & 20.34\textsubscript{±1.82} & 26.90\textsubscript{±2.41} & 25.85\textsubscript{±0.45} & 24.28\textsubscript{±2.17} \\ 
MATH & \textbf{4.60} & 3.2\textsubscript{±1.04} & 2.8\textsubscript{±0.40} & 2.33\textsubscript{±0.61} & 2.73\textsubscript{±0.23} & 3.53\textsubscript{±0.50} & 2.8\textsubscript{±0.20} \\ 
CSQA & 20.56 & \textbf{57.44\textsubscript{±1.12}}$^{\color{red}{\uparrow\scriptsize11.38\%}}$ & 55.23\textsubscript{±1.47} & 55.42\textsubscript{±0.64} & 53.65\textsubscript{±0.22} & 52.96\textsubscript{±0.29} & 51.57\textsubscript{±1.15} \\ 
OBQA & 37.80 & 57.2\textsubscript{±1.59} & 52.33\textsubscript{±0.61} & 54.87\textsubscript{±2.02} & 54.6\textsubscript{±1.04} & \textbf{57.47\textsubscript{±2.64}}$^{\color{red}{\uparrow\scriptsize9.82\%}}$ & 52.93\textsubscript{±1.81} \\ 
STQA & 54.14 & 58.85\textsubscript{±1.74} & \textbf{61.04\textsubscript{±3.06}}$^{\color{red}{\uparrow\scriptsize3.72\%}}$ & 60.58\textsubscript{±3.09} & 59.89\textsubscript{±3.13} & 59.19\textsubscript{±1.21} & 59.08\textsubscript{±2.87} \\ 
\bottomrule
\end{tabular}
}
\caption{Performance of various models at six granularity levels, including standard deviation (\textit{±std}). The best performance is boldfaced, and red text shows the relative improvement (\%) for the highest vs. lowest performance in six levels. \textit{Only Answer}: Student models are fine-tuned to directly predict answers without CoT.}
\label{tab:granularity_performance_all}
\end{table*}

\newpage
\section{Padding Procedure for Matched-Length CoT Variants} 
\label{appendix:Padding}
The following algorithm outlines the process of constructing matched-length CoT variants $\mathcal{D}_g^\prime$, ensuring that sequences from lower granularity levels are padded to match the length of higher granularity sequences. This process is designed to isolate the impact of granularity from sequence length during evaluation.




\begin{algorithm}[h]
\caption{Dynamic Padding for Matched-Length CoT Variants}
\label{alg:padding}
\KwIn{$\mathcal{D}_1, \mathcal{D}_5$ (CoT outputs for levels 1, 5)}
\KwOut{$\mathcal{D}_1^\prime$: Padded $\mathcal{D}_1$ matching $\mathcal{D}_5$}

Load $\mathcal{D}_1, \mathcal{D}_5$, ensure lengths match\;

\ForEach{$(x_1, x_5) \in (\mathcal{D}_1, \mathcal{D}_5)$}{
  Tokenize $x_1, x_5$ to get lengths $\text{len}(T_1)$ and $\text{len}(T_5)$\;
  
  \If{$\text{len}(T_1) < \text{len}(T_5)$}{
    Compute $\text{repeat\_factor} = \text{len}(T_5)/\text{len}(T_1)$\;
    Repeat and truncate $T_1$ to match $\text{len}(T_5)$, decode to $x_1^{\text{tmp}}$\;
    
    \If{"The answer is" in $x_1$}{
      Split $x_1$ as $[\text{prefix}, \text{suffix}]$ at "The answer is"\;
      $x_1^\prime = \text{prefix} + x_1^{\text{tmp}} + \text{"The answer is"} + \text{suffix}$\;
    } \Else{
      $x_1^\prime = x_1^{\text{tmp}}$\;
    }
    
    Append $x_1^\prime$ to $\mathcal{D}_1^\prime$\;
  }
}

Save $\mathcal{D}_1^\prime$ and log statistics\;
\end{algorithm}

\section{Analysis of Symbolic CoT Performance in CoT distillation} 
\label{appendix:Symbolic CoT}
To investigate the suboptimal performance of Symbolic CoT, we conducted a case study.

Question: janet’s ducks lay 16 eggs per day. she eats three for breakfast every morning and bakes muffins for her friends every day with four. she sells the remainder at the farmers' market daily for \$2 per fresh duck egg. How much in dollars does she make every day at the farmers' market?
\begin{tcolorbox}[colback=gray!10, 
  colframe=black, 
  title=\textbf{BLOOM 3B Output} ]
{\small
Let:  
\begin{align*}
D &= 16 \quad \text{(eggs laid per day)}\\
E &= 3 \quad \text{(eggs eaten per day)}\\
B &= 3 \quad \text{(breakfasts per day)}\\
M &= 4 \quad \text{(muffins per day)}\\
S &= D - E = 16 - 3 = 13 \quad \text{(sold eggs per day)}\\
R &= \text{remaining eggs after breakfasts and muffins}\\
T &= \text{total daily earnings}
\end{align*}

\noindent
1. \( \forall E: E = 3 \) (Janet eats 3 eggs per day)\\
2. \( \forall B: B = 3 \) (Janet bakes muffins for her friends every day)\\
3. \( \forall M: M = 4 \) (Janet sells the remaining eggs at the farmers' market daily for \$2 per fresh duck egg)\\
4. \( \forall R: R = D - E - M = 16 - 3 - 4 = 9 \) (remaining eggs after breakfasts and muffins)\\
5. \( \forall T: T = R \times 2 \Rightarrow T = 13 \times 2 = 26 \) (total daily earnings)\\

\noindent
The answer is \textcolor{red}{26}.
}
\end{tcolorbox}

\begin{tcolorbox}[  colback=gray!10, 
  colframe=black, title=\textbf{Gemma 2B Output}]
{\small
Let:  
\begin{align*}
E &= 16 \quad \text{(total eggs laid per day by ducks)}\\
B &= 3 \quad \text{(eggs eaten for breakfast per day)}\\
M &= 4 \quad \text{(eggs baked into muffins per day)}\\
R &= \text{remaining eggs after breakfast and muffins}\\
S &= \text{eggs sold at farmers' market per day}\\
P &= 2 \quad \text{(\$ per egg at farmers' market)}\\
T &= \text{total earnings per day at farmers' market}
\end{align*}

\noindent
1. \( \forall E: E = 16 \) (total eggs laid per day)\\
2. \( \forall B: B = 3 \) (Janet eats 3 eggs for breakfast)\\
3. \( \forall M: M = 4 \) (Janet bakes 4 muffins daily)\\
4. \text{Remaining eggs after breakfast and muffins:} 
   \[ R = E - B - M \Rightarrow R = 16 - 3 - 4 = 9 \] 
5. \( \forall S: S = R \Rightarrow S = 9 \) (Janet sells the remaining eggs)\\
6. \text{Total earnings:} 
   \[ T = S \times P \Rightarrow T = 9 \times 2 = 18 \] 

\noindent
The answer is \textcolor{darkgreen}{18}.
}
\end{tcolorbox}

The error in \textbf{BLOOM 3B} arises from unnecessary variable assignments and mismanagement of numerical values. While it correctly defines and calculates \( R = 9 \) (the remaining eggs), it later incorrectly substitutes \( R = 13 \) into the final computation, leading to an incorrect total earnings prediction of \textbf{\$26} instead of \textbf{\$18}. The symbolic expressions themselves are structurally correct, but the model fails to maintain numerical consistency throughout the reasoning process. This suggests that Symbolic CoT lacks a robust mechanism for tracking variable dependencies and verifying intermediate values, especially for weaker SLMs. We identified and summarized several possible reasons for its poor performance:

\begin{enumerate}
    \item \textbf{Task Relevance and Reasoning Depth}: According to the original paper, Symbolic CoT is primarily designed for logical reasoning tasks~\cite{xu2024faithfullogicalreasoningsymbolic}. However, our datasets focus on mathematical and commonsense reasoning, where the advantages of symbolic reasoning—particularly its effectiveness in handling deeper reasoning—do not manifest as clearly.
    \item \textbf{Implementation Differences}: The original study employed multiple stages and corresponding special tokens to enhance symbolic reasoning. In contrast, our implementation only adopted the symbolic reasoning format without these additional mechanisms, which might have impacted its effectiveness.
    \item \textbf{Pretraining Data Constraints}: SLMs have relatively limited pretraining corpora, which likely contain fewer instances of symbolic reasoning formats. As a result, weaker models struggle to acquire symbolic reasoning capabilities with only a small number of training samples.
\end{enumerate}

\newpage
\section{Student Performance across Different Teacher Models} 
\label{appendix:Teacheracc}

\begin{figure}[!htb] 
    \centering
    \begin{subfigure}[b]{0.32\columnwidth} 
        \centering
        \includegraphics[width=\columnwidth]{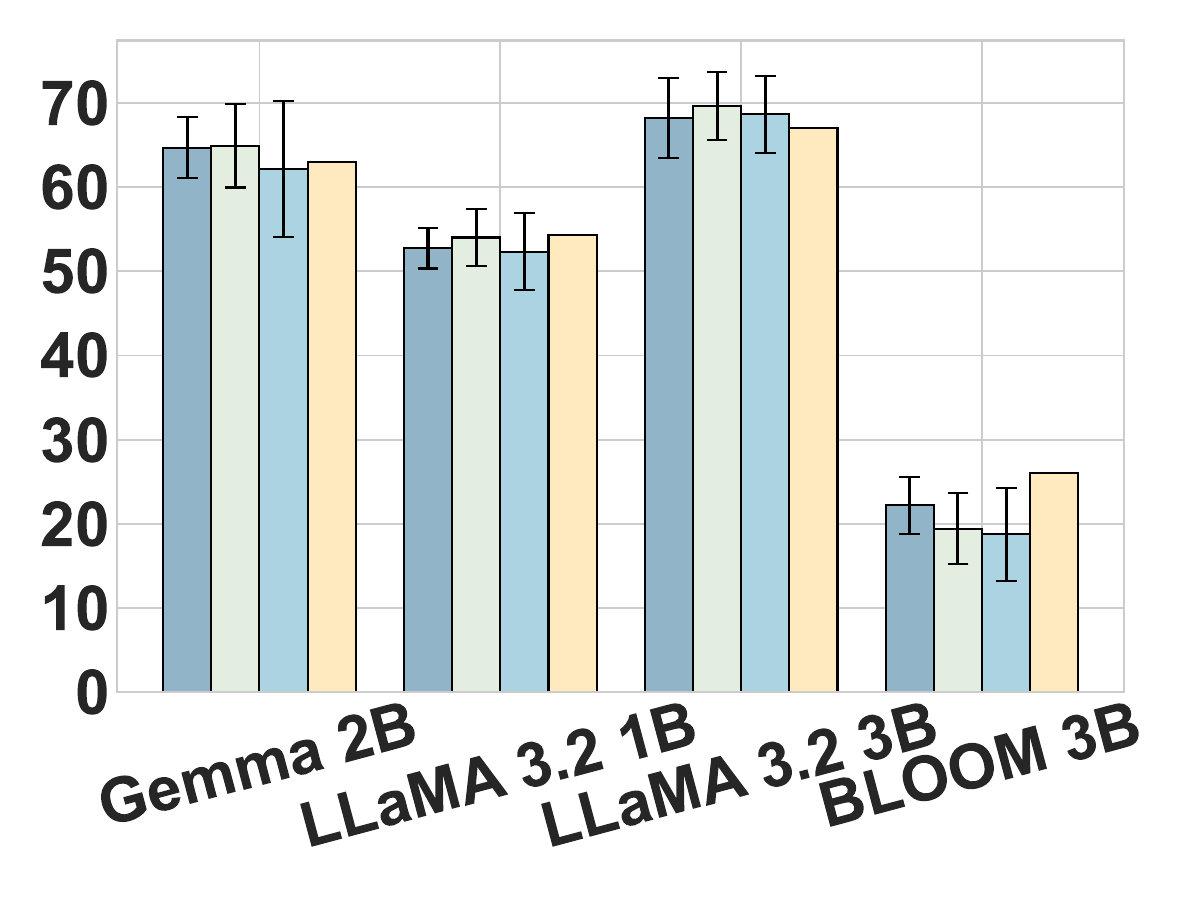}
        \caption{SVAMP}
    \end{subfigure}%
    \hfill
    \begin{subfigure}[b]{0.32\columnwidth} 
        \centering
        \includegraphics[width=\columnwidth]{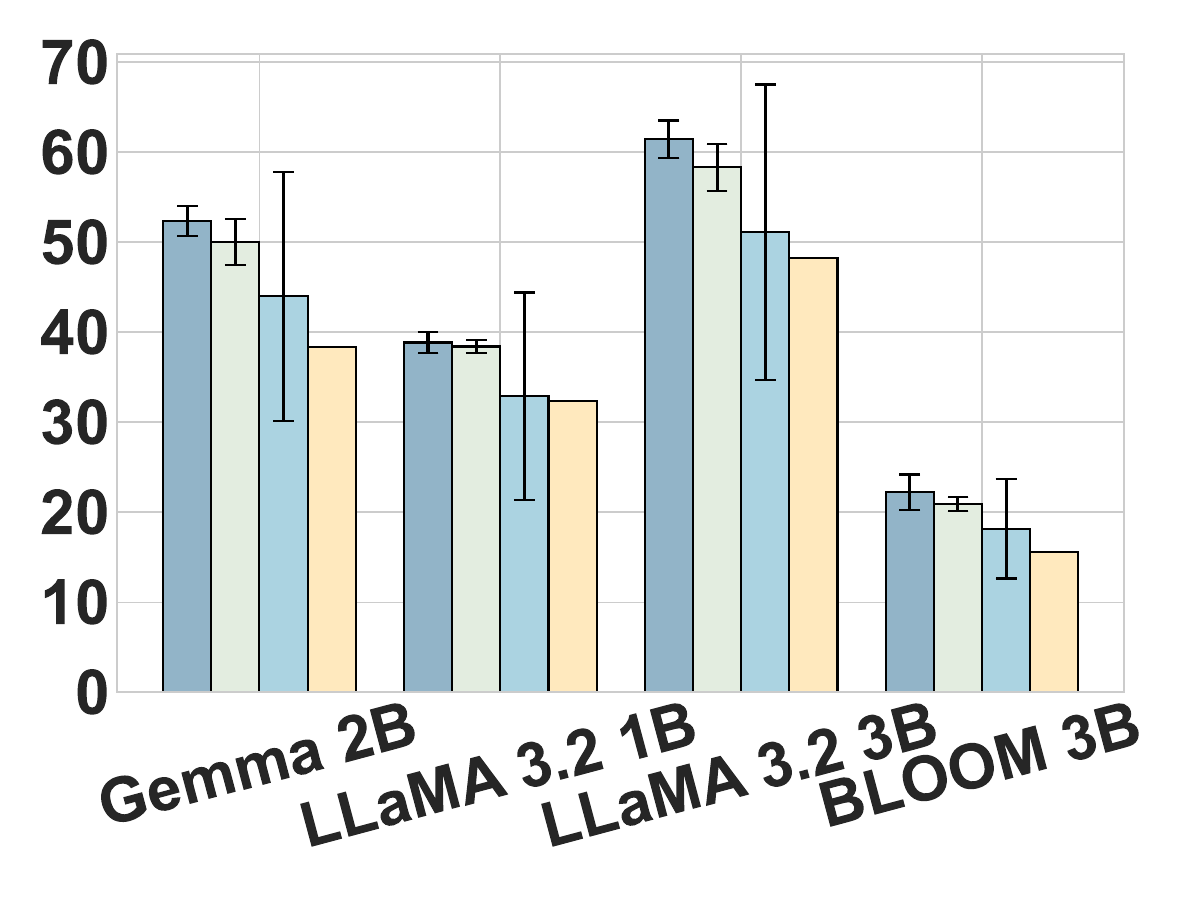}
        \caption{GSM8K}
    \end{subfigure}
    \hfill
    \begin{subfigure}[b]{0.32\columnwidth} 
        \centering
        \includegraphics[width=\columnwidth]{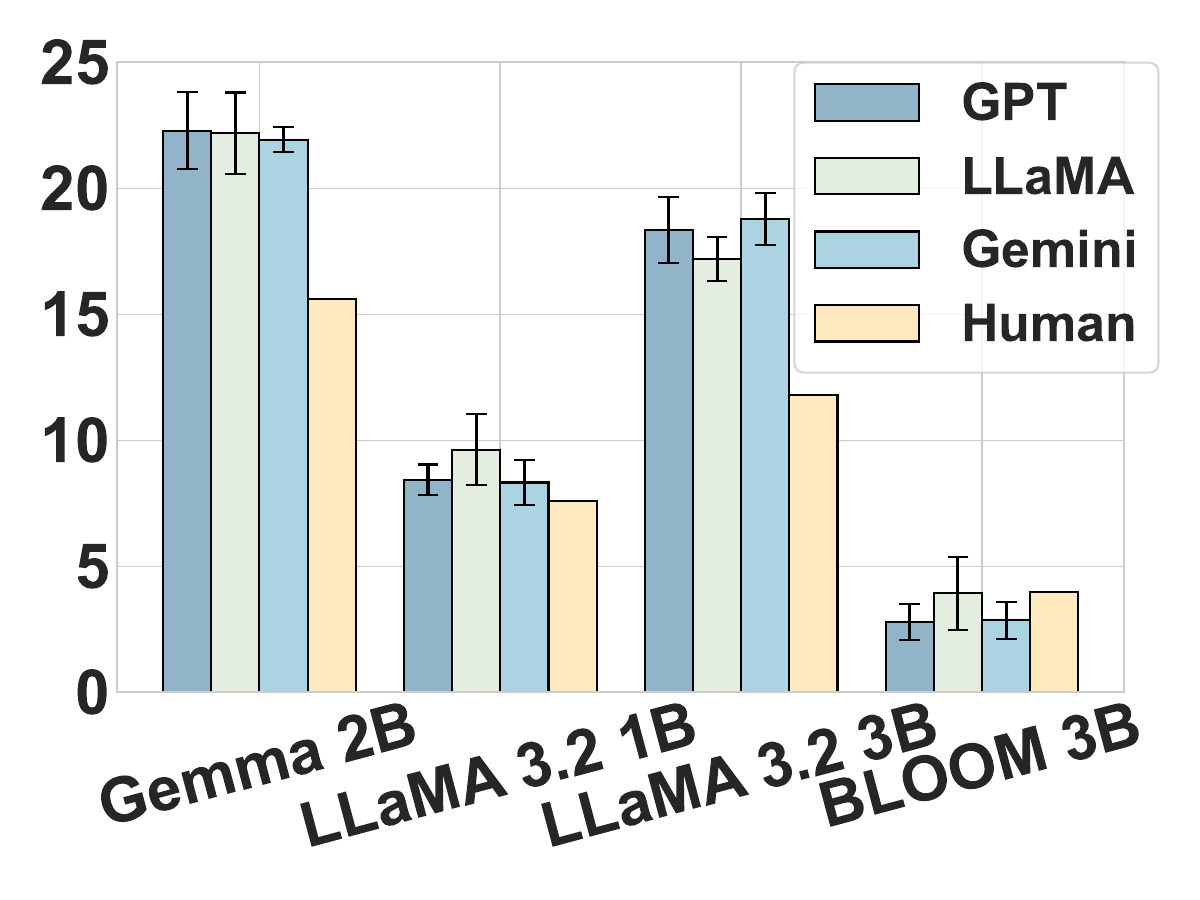}
        \caption{MATH}
    \end{subfigure}

    \begin{subfigure}[b]{0.32\columnwidth} 
        \centering
        \includegraphics[width=\columnwidth]{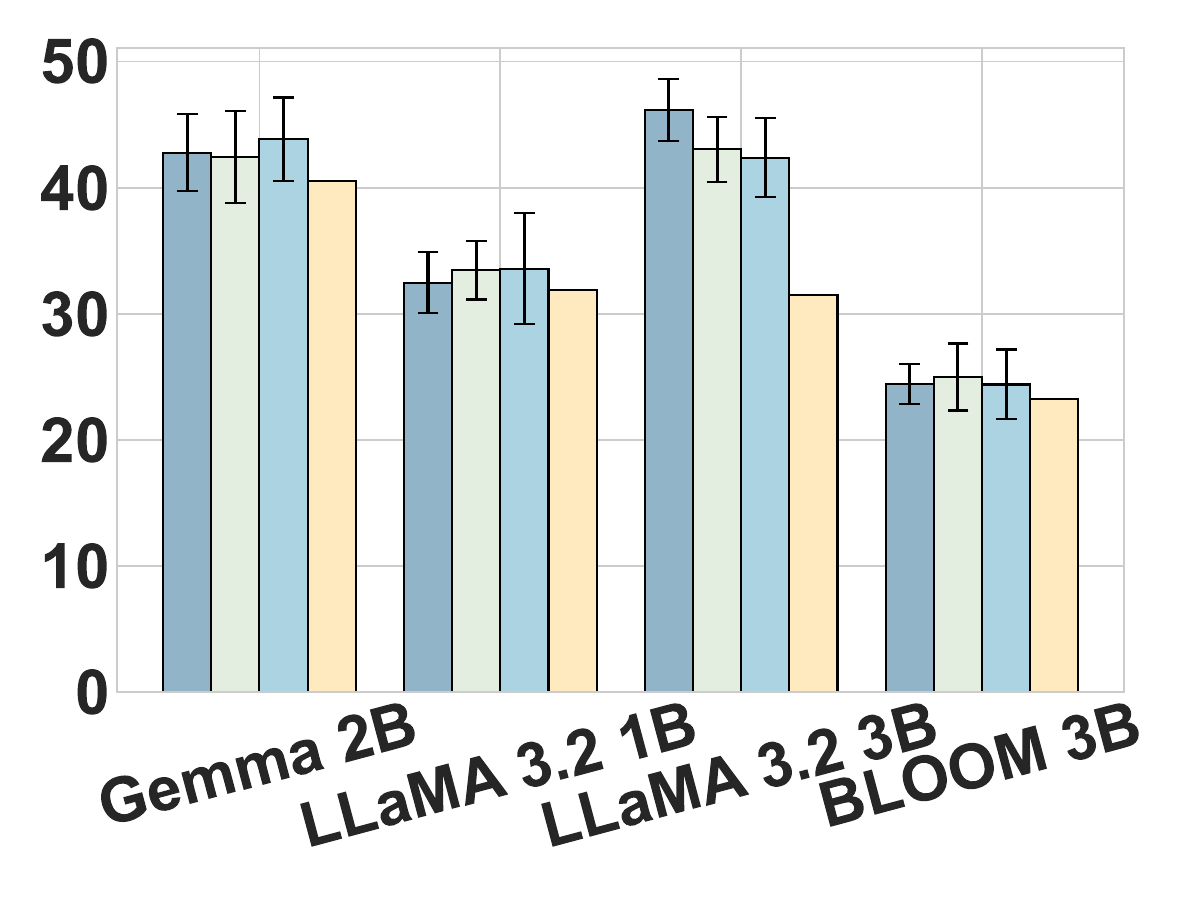}
        \caption{AQuA-RAT}
    \end{subfigure}
    \hfill
    \begin{subfigure}[b]{0.32\columnwidth} 
        \centering
        \includegraphics[width=\columnwidth]{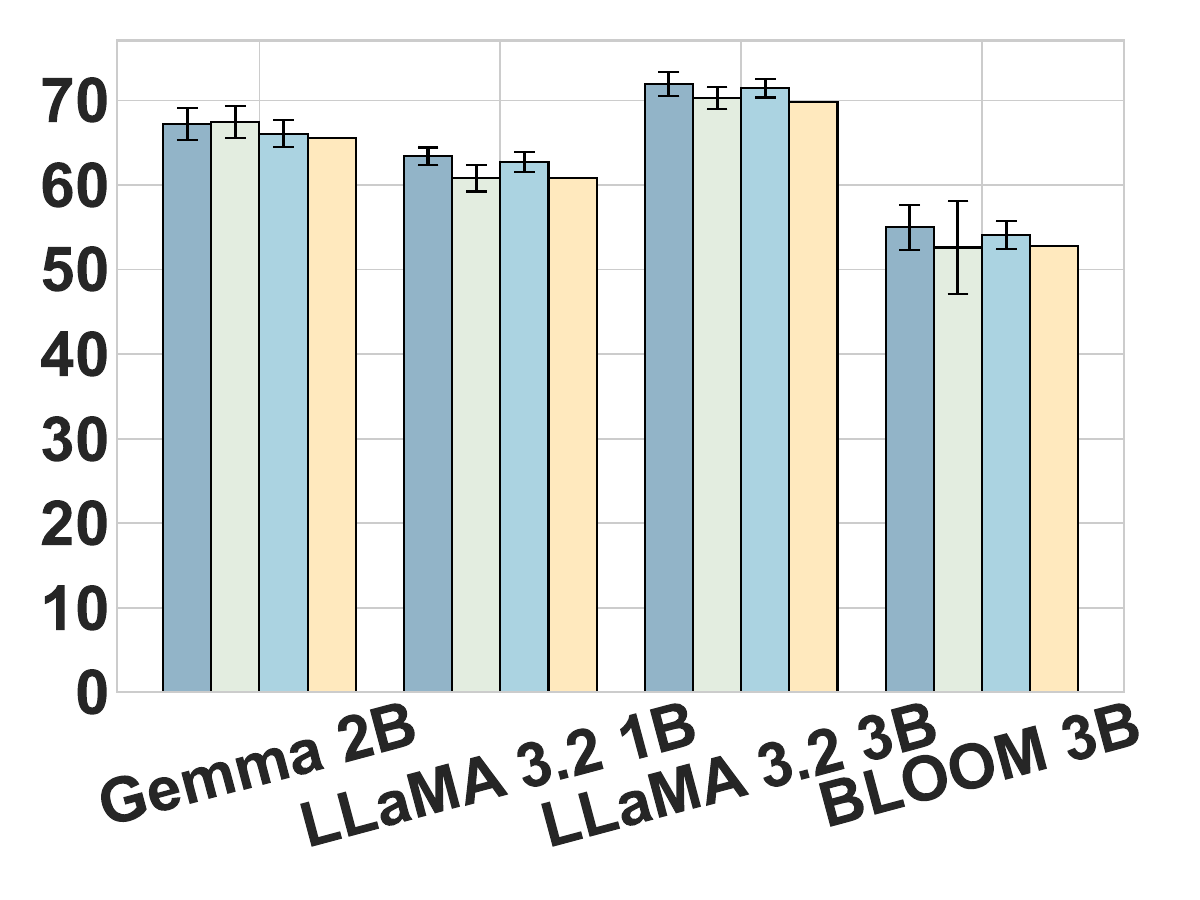}
        \caption{CommonsenseQA}
    \end{subfigure}
    \hfill
    \begin{subfigure}[b]{0.32\columnwidth} 
        \centering
        \includegraphics[width=\columnwidth]{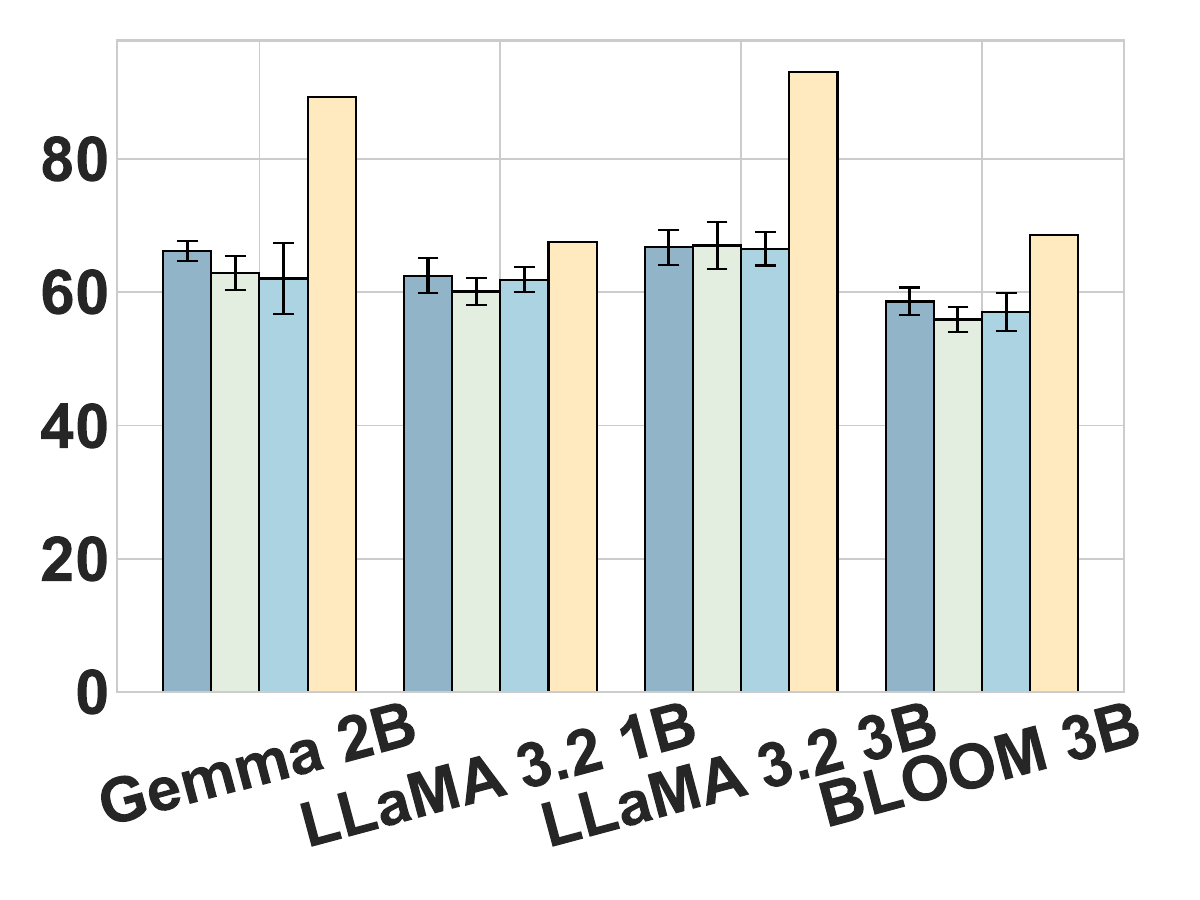}
        \caption{StrategyQA}
    \end{subfigure}

    \caption{Student model performance across different teacher models. Each bar represents the average accuracy of a specific student model trained on CoT from different teacher models.}
    \label{fig:wholeteacher}
\end{figure}

\end{document}